\RequirePackage{fix-cm}
\documentclass[smallextended]{svjour3}       
\smartqed  
\usepackage{amssymb}
\usepackage{graphicx}
\usepackage{epstopdf}
\usepackage[cmex10]{amsmath}
\usepackage{amsmath}
\usepackage{amssymb}
\usepackage{url}
\usepackage{array}
\usepackage{algorithm}
\usepackage{algpseudocode}
\usepackage{subfig}
\usepackage{tabularx}
\usepackage{wrapfig}
\usepackage{lipsum}
\usepackage{tikz}
\usepackage{xcolor}
\usepackage{lineno}
\usepackage[titletoc,title]{appendix}
\usetikzlibrary{bayesnet}
\usepackage[font={small}]{caption}

\newtheorem{bobdef}{Definition}

\algnewcommand\algorithmicforeach{\textbf{for each}}

\algdef{S}[FOR]{ForEach}[1]{\algorithmicforeach\ #1\ \algorithmicdo}
\begin{document}
\title{Bag of biterms modeling for short texts\footnote{This paper is an extension version of the PAKDD2016 Long Presentation paper “Mai, K., Mai, S., Nguyen, A., Van Linh, N.,\& Than, K. (2016). Enabling Hierarchical Dirichlet Processes to Work Better for Short Texts at Large Scale. In Advances in Knowledge Discovery and Data Mining (pp. 431-442). Springer International Publishing.}
}


\author{Anh Phan Tuan \and Bach Tran \and Thien Nguyen Huu \and Linh Ngo Van \and Khoat Than
}


\institute{
Anh Phan Tuan\textsuperscript{1}   \email{tuananhlfc@gmail.com}    
           \and \\
Bach Tran\textsuperscript{2}            \email{tranxuanbach1412@gmail.com}     
           \and \\
Thien Nguyen Huu\textsuperscript{3}            \email{thien@cs.uoregon.edu}     
           \and \\
Ngo Van Linh\textsuperscript{4}   \email{linhnv@soict.hust.edu.vn}    
           \and \\
Khoat Than \textsuperscript{5}             \email{khoattq@soict.hust.edu.vn}     
          \and \\         
\textsuperscript{1,2,4,5}  School of Information \& Communication Technology, Hanoi University of Science and Technology, No. 1, Dai Co Viet road, Hanoi, Vietnam \\
\textsuperscript{3}  Department of Computer and Information Science, University of Oregon
}

\date{Received: date / Accepted: date}
\maketitle

\begin{abstract}
Analyzing texts from social media encounters many challenges due to their unique characteristics of shortness, massiveness, and dynamic. Short texts do not provide enough context information, causing the failure of the traditional statistical models. Furthermore, many applications often face with massive and dynamic short texts, causing various computational challenges to the current batch learning algorithms. This paper presents a novel framework, namely Bag of Biterms Modeling (BBM), for modeling massive, dynamic, and short text collections. BBM comprises of two main ingredients: (1) the concept of Bag of Biterms (BoB) for representing documents, and (2) a simple way to help statistical models to include BoB. Our framework can be easily deployed for a large class of probabilistic models, and we demonstrate its usefulness with two well-known models: Latent Dirichlet Allocation (LDA) and Hierarchical Dirichlet Process (HDP). By exploiting both terms (words) and biterms (pairs of words), the major advantages of BBM are: (1) it enhances the length of the documents and makes the context more coherent by emphasizing the word connotation and co-occurrence via Bag of Biterms, (2) it inherits inference and learning algorithms from the primitive to make it straightforward to design online and streaming algorithms for short texts. Extensive experiments suggest that BBM outperforms several state-of-the-art models. We also point out that the BoB representation performs better than the traditional representations (e.g, Bag of Words, tf-idf) even for normal texts.
\keywords{short texts \and document representation \and topic modeling \and short text classification}
\end{abstract}

\section{Introduction}	\label{sec:I}
In recent years, short texts have emerged as a dominant source of text data, being used in the major activities on the web such as search queries, tweets, tags, messages and social network posts. It is therefore crucial for us to be able to automatically analyze such large amount of short texts and gain valuable knowledge from it. Conventional topic modeling techniques such as pLSA \cite{hofmann1999probabilistic}, LDA \cite{blei2003latent}, and HDP \cite{teh2006hierarchical} are the natural considerations to perform such analysis as they have been demonstrated as the successful techniques for text analysis with the usual long documents. Unfortunately, the direct application of those topic modeling techniques causes various issues for short texts due to their unique characteristics of being short, informal, massive and dynamic. A typical issue concerns the shortness of the texts. In particular, the conventional topic models exploit the word co-occurrence information in the same topics to identify the main topics of documents; however, the word co-occurrences are very infrequent (data sparsity) for short texts. This has led to the poor performance of topic models for short texts despite the fact that the text collection might be large \cite{tang2014understanding,than2014dual}. The challenges caused by short texts can also be found in the other prior studies \cite{sahami2006web,bollegala2007measuring,yih2007improving,banerjee2007clustering,schonhofen2009identifying,phan2008learning,mehrotra2013improving,grant2011online,chengxu2014,hong2010empirical}.

\par
In order to alleviate the problem of data sparsity for short texts, three major heuristic strategies have been proposed, i.e, (1) the strategy \cite{banerjee2007clustering,schonhofen2009identifying,phan2008learning,quiang2017topic,zhao2017word,li2017enhancing} that employs external resources or human knowledge to overcome the challenge of lacking information, (2) the strategy \cite{weng2010twitterrank,hong2010empirical,mehrotra2013improving,jiang2016biterm,bicalho2017general} that modifies input data to enhance the word co-occurrence information and then simply pass the new input to the conventional topic models, and (3) the strategy \cite{yang2018topic,zuo2016topic,quan2015short,yan2014btm} that develops new probabilistic topic models better suited for with short texts. Obviously, the first strategy is an effective approach to enrich short texts. Unfortunately, to the best of our knowledge, all existing works only focus on batch learning, and as a consequence cannot be considered as conventional methods to deal with short texts that are massive and dynamic in practice. Regarding the second strategy, the popular technique is to combine multiple documents into a single one. This technique is simple and beneficial as it artificially increases the length of the input documents and enhances the word co-occurrence information. However, a fatal issue of this strategy is that it has to know the metadata of the data source in advance to guide the combination process. Consequently, the second strategy is also not universal as it cannot be used for all scenarios in practice (especially the ones without metadata). Finally, the third strategy, i.e, developing new statistical models, seems to be the most flexible approach among the three aforementioned strategies to deal with the statistical issues in data very well. However, designing better models is not always an easy effort for different application domains. 

\par
In this paper, we propose a hybrid strategy to address the sparsity problem of short texts combining the second and the third strategies mentioned above (i.e, both modifying the input data and developing new topic models for short texts). Our work also demonstrates the massive and dynamic challenges of short texts (which make batch learning inefficient) by focusing on the online and streaming methods. In particular, we present a framework for short texts with two major ingredients. First, BBM represents each document from corpus by Bag of biterms (BoB). A BoB is a bag that contains both terms and biterms appearing in a text snippet. A biterm, in turn, is a pair of terms in the text \cite{yan2014btm}. The employment of both terms and biterms in BoB offers several benefits for short text, i.e, (i) it naturally lengthens the document representation and thus helps to reduce the negative effects of shortness \cite{tang2014understanding,than2014dual}, (ii) it helps to emphasize the co-occurrence of words by directly including the biterms in the modeling, and (iii) it also draws attention to connotations of the terms themselves, rather than just focusing on the word co-occurence revealed by biterms.

Second, BBM explitcitly models both terms and biterms in BoB. On one side, the framework inherits the way to encode each term in documents from the previous models. On the other side, we utilize the assumption that the two words in a biterm share the same topic and then model each word seperately. Therefore, BBM is essentially different from the previous works that have only attempted to lengthen the input documents (i.e, via external resources or multiple document aggregation) at the preprocessing step. The advantages of BBM include: (i) it does not assume the availability of any additional information or resources (e.g, the metadata information to guide the document aggregation in the previous methods), thus being applicable to a wide ranges of existing topic models (base models) to enable them to work well with short texts, (ii) the specific models built on BBM inherit the full advantages of the base models (e.g, easily designing online and streaming algorithms as in \cite{wang2011online,broderick2013streaming,duc2017keeping} and exploiting human knowledge in streaming environment as in \cite{duc2017keeping}).


\par
We conduct extensive experiments to demonstrate the benefits of the proposed framework. First, we demonstrate the advantages of BoB over the traditional schema to represent documents (i.e, Bag of Words, tf-idf) in both the supervised and unsupervised learning scenarios. Second, we find that the implementations of BBM outperform the premitives even when BoB representation is applied. Finally, BBM exceeds the state-of-the-art performance for short texts \cite{yan2014btm,broderick2013streaming,duc2017keeping} in both online and streaming settings. 

\par 
The rest of the paper is organized as follows. Section \ref{sec:II} briefly reviews the previous approaches to deal with short texts and the background necessary for this paper. Section \ref{sec:III} introduces the mechanism of BBM framework via the concept of BoB representation and the way to include BoB in statistical models. Section \ref{sec:IV} presents two particular case studies when applying BBM to probabilistic models. Section \ref{sec:V} describes the experiments and the evaluation results of the proposed techniques. Finally, some conclusions are made in Section \ref{sec:VI}.
\section{Related work and Background}\label{sec:II}
\subsection{Topic models on Short Texts}
Due to the importance of short texts, much research effort has been spent on developing analysis methods for this kind of text. In this section, we focus on the methods that are employed in the previous research for topic models. These methods can be classified into three major groups as mentioned in Section \ref{sec:I}.

The first strategy exploits the external resources or human knowledge to deal with short texts. Among a variety of external resources, perhaps Wikipedia is the most conventional resource in the prior research. The content from Wikipedia is often used to generate additional information/text to augment the short text input \cite{banerjee2007clustering,schonhofen2009identifying,phan2008learning}. More recently, word embeddings have been widely adopted as an effective external knowledge to assist the models for short texts. In particular, the embedding-based topic model (ETM) in \cite{quiang2017topic} builds word embeddings from external sources to generate semantic knowledge about the documents on which the aggregation decisions for the documents with short texts are relied. Afterward, ETM utilizes a Markov random field regularized model to extract latent topics from the generated pseudo documents. In \cite{zhao2017word}, the authors present the word  embedding informed focused topic model (WEI-FTM) that employs word embeddings as a prior knowledge for the distributions of the topics. While LDA considers each topic as a distribution over all words, WEI-FTM allows each topic to focus on fewer words. Finally, the GPU Dirichlet Multinomial Mixture (GPU-DMM) method in \cite{li2017enhancing} exploits the general semantic word relations during the topic inference process. These semantic relations are learned from external resources, and then incorporated into the generalized Polya urn (GPU) model \cite{mimno2011optimizing} in the sampling process. A common weakness of these strategies is that there would be no guarantee about the correlation between the universal corpus and the short text corpora since most of the short texts are produced from social networks with informal contents and noises. Moreover, they cannot work when data arrives continuously. Our work shows that BBM can also exploit word embeddings as a prior to improve experimental results in streaming environment.

\par
In the second strategy to handle short texts, the input data is modified in rational ways whose results are fed into conventional topic models to induce topics. Specifically, Weng et al. \cite{weng2010twitterrank}, Hong et al. \cite{hong2010empirical} and Mehrotra et al. \cite{mehrotra2013improving} aggregate tweets that share the users, the words and the hashtags respectively. In \cite{quan2015short}, the authors postulate that each short text snippet is sampled from an unobserved long pseudo-document. Based on this idea, the self-aggregation based topic model (SATM) \cite{quan2015short} is proposed to generate a set of $D$ regular-sized documents and then use each document to obtain a few short texts for topic models. \cite{jiang2016biterm} presents a novel word co-occurrence network to construct pseudo documents. This network is called the biterm pseudo document topic model (BPDTM) and exploits the triangle relations among words (i.e, three words are semantically close if every two of them co-occur in the corpus). Finally, in \cite{bicalho2017general}, the authors identify similar components (e.g. words or bi-grams) from external corpora to enrich the original short text documents. These methods share the same issue that the aggregating strategies cannot be applied to a wide range of datasets since they need to know the metadata of data sources (e.g, the hashtags, users etc.). Therefore, these methods might be only applicable for specific datasets but unsuitable for others. In addition, even if the aggregating strategies are reasonable, it is likely that the generated pseudo documents become more ambiguous due to the noises or ambivalent words. In this work, we also attempt to modify the input text to generate a new representation of the short texts. However, instead of explicitly aggregating the text snippets at the preprocessing step with metadata and feeding the new pseudo documents into the topic models, we generate additional clues based on only the words of the short text input and directly model those clues in the modeling step. This helps the proposed method avoid the metadata (thus being applicable to a wide range of datasets) and better exploit the additional clues via deeper modeling and analysis.

Finally, in the third strategy, new probabilistic topic models with better modeling mechanisms are proposed to circumvent the problems of short texts. The co-occurring document topic model (COTM) in \cite{yang2018topic} assumes that each document has a probability distribution over two set of topics (i.e, the formal topics and the informal topics). A switch variable is then introduced to determine the set of topics to which the text snippet belongs. The pseudo-document-based topic model (PTM) and the sparsity-enhanced PTM model (SPTM) in \cite{zuo2016topic}, on the other hand, consider each short text as belonging to one and only one pseudo document, and then utilize the self-aggregation topic model (SATM) \cite{quan2015short} to generate these pseudo documents. Although these new models obtain better results than the conventional topic models, almost all of the current proposed models cannot be applied directly to raw input data. They have to use some heuristic methods to enrich the input data before the new modeling technique can be applied. Consequently, these models still have the limitation of the first and the second strategies to some extent.  Perhaps the most related study to our work in this paper is \cite{yan2013biterm,yan2014btm} that also employs biterms in the novel biterm topic model (BTM), currently a strong baseline for analyzing short texts. In BTM, the model generates all pairs of words (called biterms) for each document and aggregate the biterms of all the documents into one collection. This is then followed by modeling the generative process for this collection. One problem of BTM is that aggregating all the documents in a corpus can make the context ambiguous, especially when the corpus involves a mixture of topics. Furthermore, this approach does not model the generative process for each document explicitly. The authors instead use a heuristic to infer the topic proportions for each document, causing the potential inconsistency between the training phase and the inference phase. In contrast, our proposed framework ameliorates the word co-occurence information by keeping the bag of biterms (BoB) within each document. We would then explicitly model the generative process for each document in the corpus using both the words and the biterms in the documents.

N-gram is also a popular representation method to enrich context information [31-33] for documents.  Some topic models such as bigram topic model [33] and topical N-grams [31] employ bigram to achieve better performance.  However, N-gram seems to be only suitable for long texts. There are some reasons why topic models using N-gram might not be effective to deal with short texts. First, since the frequency of N-grams in short-text corpus is small (i.e., almost bigrams occur in one or two documents), it is difficult for these models to learn hidden topics. Second, they do not utilize efficiently co-occurrence in short texts because they just model consecutive words. A short snippet often includes a topic and therefore all co-occurring words in this snippet should be paired to gain better understanding. Finally, they do not use both unigram and bigram to enrich the representation of short text.  While bigram topic model [33] only learns hidden topics on bigram, topical N-grams [31] uses a mechanism to model either unigram or bigram at a time.

\subsection{Background}
Latent Dirichlet Allocation (LDA) \cite{blei2003latent} and Hierarchical Dirichlet Process (HDP)\cite{teh2006hierarchical} are the most popular topic models for regular texts.  However, they cannot deal with short texts \cite{tang2014understanding,than2014dual,mai2016enabling}. This subsection briefly describe them, serving as the base models for BBM in the next two case studies.
\subsubsection{Overview of Latent Dirichlet Allocation (LDA).}

\begin{figure}
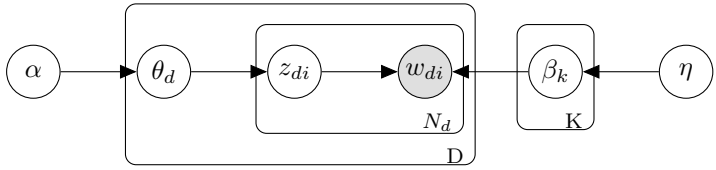

\centering
      \tikz{ %
        \node[latent] (alpha) {$\alpha$} ; %
        \node[latent, right=of alpha] (theta) {$\theta_d$} ; %
        \node[latent, right=of theta] (z) {$z_{di}$} ; %
        \node[obs, right=of z] (w) {$w_{di}$} ; %
        \node[latent, right=of w] (beta) {$\beta_k$} ; %
        \node[latent,right=of beta] (eta) {$\eta$};%
        \plate[inner sep=0.14cm, xshift=0cm, yshift=0.12cm] {plate1} {(z) (w)} {$N_d$}; %
        \plate[inner sep=0.14cm, xshift=0cm, yshift=0.12cm] {plate2} {(theta) (plate1)} {D}; %
        \plate[inner sep=0.14cm, xshift=0cm, yshift=0.12cm] {plate3} {(beta)} {K};
        \edge {alpha} {theta} ; %
        \edge {theta} {z} ; %
        \edge {z,beta} {w} ; %
        \edge {eta} {beta};
      }
	\caption{The graphical representation of Latent Dirichlet Allocation (LDA)}
	\label{fig:graph_lda_model}
\end{figure}
\par 
Suppose that the dataset contains $D$ documents and $K$ topics, and each document contain $N$ words. A hyperparameter $\alpha$ contributes the distribution of topic mixture $\theta$, and $\beta$ is the topic distribution over $V$ words of the vocabulary. The graphical representation of LDA is shown in Figure \ref{fig:graph_lda_model}. The generative process of LDA is as follows: 
\newpage
\begin{enumerate}
\item Draw topics $ \beta_{k} \sim \text{Dirichlet}(\eta) $ for $ k \in [1, K] $
\item For each document $ d \in [1,D] $:
	\begin{enumerate}
		\item Draw topic proportions $ \theta_{d} \sim \text{Dirichlet}(\alpha) $
		\item For each word $w_{dn} \in \{1,...,N_d\}$: 
		\begin{enumerate}
			\item Draw topic assignment $z_{dn} \sim \text{Multinomial}(\theta_d) $
			\item Draw word $w_{dn} \sim \text{Multinomial}(\beta_{z_{dn}})$
		\end{enumerate}
	\end{enumerate}
\end{enumerate}

There are several effective algorithms to enable LDA deal with massive and dynamic data, i.e, the online LDA  \cite{hoffman2010online}, the streaming variation bayes (SVB) \cite{broderick2013streaming}, and the method to exploit human knowledge in streaming environment \cite{duc2017keeping}. 

\subsubsection{Overview of Hierarchical Dirichlet Process (HDP).}
\par 
\begin{figure}
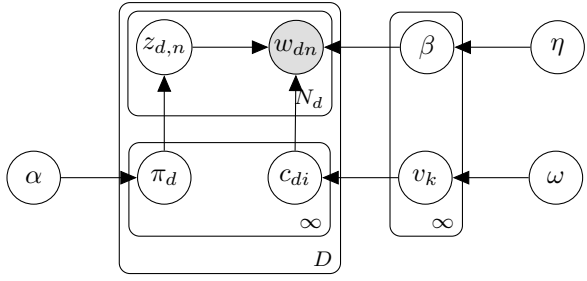

      \centering
      \tikz{ %
      	\node[latent] (alpha) {$\alpha$}; %
      	\node[latent, right = of alpha] (pi) {$\pi_d$}; %
      	\node[latent, above = of pi] (zDN) {$z_{d,n}$};
      	\node[obs, right = of zDN] (w0) {$w_{dn}$};
      	\node[latent, right = of pi] (cDI) {$c_{di}$}; %
      	\node[latent, right = of cDI] (vK) {$v_k$}; %
      	\node[latent, right = of vK] (omega) {$\omega$};
      	\node[latent, right = of w0] (beta) {$\beta$};
      	\node[latent, right = of beta] (eta) {$\eta$};
      	\edge {eta} {beta}; %
      	\edge {vK} {cDI}; %
      	\edge {omega} {vK};
      	\edge {beta, cDI} {w0};
      	\edge {zDN} {w0};
      	\edge {pi} {zDN};
      	\edge {alpha} {pi};
      	\plate {uni}{(zDN)(w0)}{$N_d$}; %
      	\plate {inf1}{(pi)(cDI)}{$\infty$};
      	\plate {}{(beta)(vK)}{$\infty$};
      	\plate {}{(uni)(inf1)}{$D$}; %
      	
     }
     \caption{The graphical representation of Hierarchical Dirichlet Process(HDP)}
     \label{fig:graph_hdp_model}
\end{figure}

While LDA is a powerful tool for discovering latent structures in texts, the fixed number of topics ($K$) makes it less flexible to handle the dynamic data. This has led to the development of Hierarchical Dirichlet Process (HDP) (shown in Figure \ref{fig:graph_hdp_model}) \cite{teh2006hierarchical}, a nonparametric variant of LDA that allows the number of topics $K$ to be countably infinite. In this paper, we follow \cite{wang2011online} to use the stick-breaking construction to have a closed-form coordinate ascent variational inference. The generative process of HDP is:

\begin{enumerate}
\item Draw an infinite number of topics, $ \beta_k \sim \text{Dirichlet}(\eta) $ for $ k \in \lbrace 1,2,3,...\rbrace $
\item Draw corpus breaking proportions, $ v_k \sim \text{Beta}(1, \omega) $ for  $ k \in \lbrace 1,2,3,...\rbrace $
\item For each document $d$:
		\begin{enumerate}
			\item Draw document-level topic indices, $c_{d,i} \sim \text{Multinomial}(\sigma(v)) $ for  $ i \in \lbrace 1,2,3,...\rbrace $
			\item Draw document breaking proportions, $ \pi_{d,i} \sim \text{Beta}(1, \alpha) $ for  $ i \in \lbrace 1,2,3,...\rbrace $
			\item For each word $w_n$:
			\begin{enumerate}
				\item Draw topic assignment $ z_{d,n} \sim \text{Multinomial}(\sigma(\pi_d)) $
				\item Draw word $w_{n} \sim \text{Multinomial}(\beta_{c_{d, z_{d,n}}})$ 
			\end{enumerate}
		\end{enumerate}
\end{enumerate}

Approximating the posterior distributions of latent variables in HDP is based on the idea of mean-field variational inference or Gibbs sampling. The details to perform inference and learning for online HDP are described in \cite{hoffman2013stochastic}.
\subsubsection{Representing documents by Bag of words (BoW)}
Bag of words has been the conventional representation of documents in text mining. In this representation, the order of the words in a document is often ignored (i.e, the exchangeability of words) and a score is often associated with a word using some weighting mechanism (e.g., \textit{tf-idf}). PLSA \cite{hofmann1999probabilistic}, LDA \cite{blei2003latent}  and HDP \cite{teh2006hierarchical} are among the typical statistical models that employ BoW as their standard representation for evaluation and comparison. However, despite the successes in analyzing regular texts, BoW exhibits many limitations in modeling short texts that have been the motivations to devise new models/methods to process this kind of text (as reviewed in Section \ref{sec:II}). In the following section, we present bag of biterms, a novel document representation to overcome the limitations of BoW for topic models with short texts.
\section{Bag of Biterms Modeling (BBM) Framework}\label{sec:III}
This section introduces the mechanism of BBM and describes several properties of the framework. BBM includes two main components: the Bag of Biterms (BoB) for representing documents and the way to model BoB in statistical models. 	
\subsection{Representing documents by Bag of biterms (BoB)}
\subsubsection{Definitions}
\label{subsec:definition-biterm}
The following two definitions about biterms and bag of biterms (BoB) are necessary for our coming discussion.
\begin{bobdef}
Given a corpus, a \textbf{biterm} is defined as a pair of words that co-occur in some documents in the corpus. For example, assume that there are two words ``character" and ``story" in some documents. The set of biterms for these documents would then includes two pairs of words: (``character", ``story") and (``story", ``character"). The definition of biterms can also be extended to cover the single terms in the document by considering each term as a biterm with two similar words (e.g., (``character", ``character") and (``story'', ``story'')). This is the main reason for the name bag of biterms since each element in the bag can be seen as a biterm.  
\end{bobdef}

\begin{bobdef}
Given a document $d$ containing $n$ distinct words $\{w_1,w_2, \ldots ,w_n\}$ with the respective frequencies $\{f_1,f_2, \ldots , f_n \}$. For \textbf{BoB}, $d$ is represented by a set of $n$ terms $w_i $ and $n(n-1)$ biterms $\tilde{w}_k = (w_{i}, w_j)$ ($i, j \in \{1,2, \ldots,n \}, j \neq i$) with the respective frequencies as follows:
\begin{itemize}
\item The frequency of $w_{i}$ is $f_i$
\item The frequency of $\tilde{w}_k = (w_{i}, w_j)$ is $\min(f_i, f_j)$ 
\end{itemize}
\end{bobdef}
 For example, assume that $d$ has three distinct words $\{w_1,w_2,w_3\}$ with the respective frequencies of $\{2,2,4\}$. In BoB, this document is represented by a BoB including $\{w_{1}, w_{2} ,$ $ w_{3}, \tilde{w}_{1} = (w_1, w_2), \tilde{w}_{2} = (w_1, w_3), \tilde{w}_{3} = (w_2, w_1), \tilde{w}_{4} = (w_2, w_3), \tilde{w}_{5} = (w_3, w_1), \tilde{w}_{6} = (w_3, w_2)\}$. The respective set of frequencies is $\{2, 2, 4, 2, 2, 2, 2, 2, 2\}$. 
 \par
Note that we choose the function \textit{min} to compute the biterm frequencies instead of the other functions such as \textit{max} or \textit{mean}. Our intuition is that the biterm $(w_i, w_j)$ is not as important as the original terms/words $w_i$ and $w_j$ so its weight should not exceed the weights of the original terms/words.

\par
The definition of BoB above is based on the underlying document representation of BoW with terms/words as the basic units/features and frequencies as their weights. In general, we can extend this definition of BoB to span other underlying document representations or weighting schemas (e.g, \textit{tf-idf}). The main idea is to employ the concept of features in the document representations, replace the words/terms with the feature values, and replace the term frequencies with the feature weights. More concretely, given a document $d$ represented by $n$ features with the feature values of $\{k_1, k_2, \ldots, k_n\}$ and the respective feature weights of $\{v_1, v_2, \ldots, v_n \}$, the BoB representation of $d$ with this underlying representation would then involve the set of biterms $\{b_{ij}=(k_i, k_j): i = 1,2, \ldots,n, j=1,2, \ldots, n\}$ and the respective set of weights $\{v_{ij}=\min(v_i, v_j): i = 1,2, \ldots,n, j=1,2, \ldots, n\}$. 

\par
Essentially, BoB can be seen as a method to transform the input data representations that preserves the format of the underlying representation. This property allows BoB to be applicable to any topic models that can work with the underlying representations, thus making BoB a model-independent representation for topic models. In addition, BoB is a context-independent representation as it only uses the terms and biterms appearing in corpus. Different from most of the current methods to modify the input representation for short texts, BoB does not require any additional information (e.g., metadata or relevant external sources) to perform its transformation. Consequently, BoB can be used as a general approach for various types of short-text datasets. 

\subsubsection{Benefits of BoB for document representation}
\label{sub:explanation} 
Although we presented BoB as an ingredient of BBM, it might also be regarded as a novel method for text representation. BoB gains various advantages over some conventional representation methods, especially in the context of short texts. In Section \ref{sec:V}, we compare efficiency of BoB to the performance of BoW when applying to both supervised and unsupervised learning algorithms. However, the obvious drawback of using BoB as a text representation method is constructing vocabulary. In theory, the number of possible terms and biterms might be up to $V_b = \frac{V(V-1)}{2} + V= \frac{V(V+1)}{2}$ where $V$ is the number of distinct words in the corpus which might be up to several hundreds of thousands. This makes BoB cost much time and memory in some applications that require the computation of vectors with dimensionality being equal to the vocabulary size. One possible solution for this issue is to restrict the number of biterms in the models using some threshold for the frequencies of the biterms in the whole corpus. The experiments show how threshold number attenuate the impact of vocabulary size. Otherwise, we can explicitly model BoB in the modeling process to avoid this problem, as we present in the next phase of BBM framework. In what follows we address some benefits when using BoB in topic models instead of conventional representation schemes. 
\par 
First, BoB reduces the negative effects of shortness of short texts. In particular, the length of the document representations in BoB is much longer than that in BoW or the underlying representation due to the addition of the biterms $b_{ij}$. In a recent research \cite{tang2014understanding}, the authors show that the document length is an important factor in such statistical models as LDA. When the document length is extremely short, the topic models are expected to have poor performance no matter how large the input corpus is. This fact is also supported by the experiments in \cite{grant2011online,mehrotra2013improving,hong2010empirical,chengxu2014} that improve the lengths of the short texts by aggregating short documents in rational ways, yielding better performance for the topic models. 

\par
Second, BoB helps to emphasize the co-occurrence of words by treating the co-occurred words as units (e.g, the biterms $b_{ij}$ with the words $w_i$ and $w_j$ in the documents). This co-occurrence modeling has been shown to be beneficial in the previous research as it helps to improve the performance of the topic models for short texts \cite{yan2013biterm}.

\par
Third, BoB can be seen as a kind of enrichment to the original document in a rational way. In particular, the biterms $(w_i,w_i)$ can be treated as the original word $w_i$ while the biterms $(w_i,w_j)$ with $i \neq j$ are the additional ``ingredients``. In a short document, the context is often ambiguous and unclear due to the lack of information. The additional ingredients (i.e, the biterms) from BoB would reinforce the context, making it clearer.

\subsection{Modeling Bag of Biterms (BoB)}
\begin{figure}[ht]
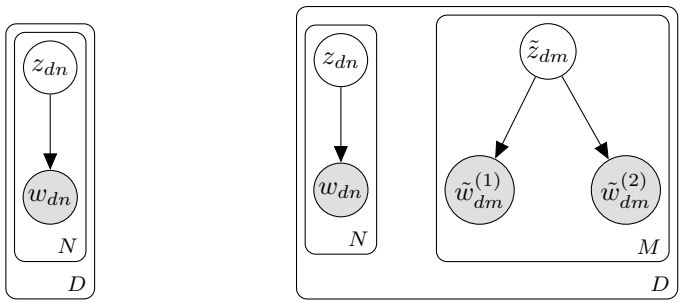

\begin{minipage}[b]{0.45\linewidth}
	\centering
      \tikz{ %
      	\node[obs] (wN) {$w_{dn}$}; %
      	\node[latent, above=of wN] (zN) {$z_{dn}$}; %
      	\edge {zN} {wN}; %
      	\plate {uni}{(zN) (wN)}{$N$}; %
      	\plate {}{(uni)}{$D$};
     }
\end{minipage}
\hspace{0.2cm}
\begin{minipage}[b]{0.45\linewidth}
	\centering
      \tikz{ %
      	\node[obs] (wN) {$w_{dn}$}; %
      	\node[latent, above=of wN] (zN) {$z_{dn}$}; %
      	\node[obs, right = of wN] (wM1) {$\tilde{w}_{dm}^{(1)}$};
      	\node[obs, right = of wM1] (wM2) {$\tilde{w}_{dm}^{(2)}$};
      	\node[latent, above=of wM1, xshift = 0.9cm] (zM) {$\tilde{z}_{dm}$}; %
      	\edge {zN} {wN}; %
      	\edge {zM} {wM1,wM2};
      	\plate {uni}{(zN) (wN)}{$N$}; %
      	\plate {bi}{(zM) (wM1) (wM2)}{$M$};
      	\plate {}{(uni)(bi)}{$D$};
     }
\end{minipage}
\caption{Modeling BoW (left) and BoB (right)}
\label{fig:modeling_process}
\end{figure}
\par
In this section, we explore how BBM explicitly models BoB in the modeling process. Figure \ref{fig:modeling_process} illustrates the distinction of modeling process between BoW (left) and BoB (right). The key idea is to use additional latent variables to generate the biterms explicitly besides the variables to generate terms in the primitive models. The generation of terms would then follow the mechanisms in the primitive topic models (i.e, LDA, HDP, etc.). In BBM, each biterm is generated by dividing it into two distinct words to be generated in the similar way as in the term part. The appealing point here is that despite separating the biterms into two terms, we only use one variable $\tilde{z}_{dm} $ to denote the topic assignment of the biterms $\tilde{w}_{dm} $ as well as their  two corresponding terms $\tilde{w}_{dm}^{(1)} $ and $\tilde{w}_{dm}^{(2)} $. Intuitively, this implies our assumption that the two words in a biterm share the same topic. For medium and long documents, this assumption might suffers from overfitting (we can avoid this problem by using bigrams instead of biterms in regular texts). However, for short text datasets, each document usually contains only one sentence with short length, leading to the tendency of the two top words to share the same topic. Since we are generating the two words of a biterm separately, it is unnecessary to construct a large vocabulary for biterms like in BoB. Consequently, the limitation on time and memory of BoB is diminished in BBM. In the next section, we will show two implementation of BBM in two probabilistic models: LDA and HDP, and we provide the generative process as well as learning algorithm of these case studies. 

\subsection{BBM Remark}
We present several advantages of BBM and explain the main reasons why it is good.
\par 
First, BBM inherits all of the advantages of the BoB representation as well as eliminates its drawbacks of requiring much time and memory. As we have described, BBM explores the topic distribution of both terms and biterms during the learning process via the BoB representation. Therefore, it possesses all the benefits of BoB such as increasing the documents length, emphasizing the word co-occurence and enriching the context of the original documents (i.e, in Section \ref{sub:explanation}). In addition, as BBM generates each biterm by dividing it into two distinct terms, we can avoid the expensive computation that involves the large vocabulary for BoB, leading to the overall reduction of time and memory.

\par
Second, BBM is a general framework that can be incorporated with any topic models to generate a new branch of models based on BBM. The topic models in this incorporation with BBM are called the primitive (base) topic models. In BBM, the inference for the hidden variables of the word and biterm parts can be simply done in a very similar way to those in the the premitive models. Moreover, it is straightforward to develop online and streaming algorithms based on the previous work \cite{hoffman2010online,broderick2013streaming,duc2017keeping}, and exploit human knowledge \cite{duc2017keeping} in the streaming environments.
\par 
Finally, BBM not only exploits the full advantages of BTM \cite{yan2014btm} but also overcomes its inherent disadvantages. By encoding the documents separately, the topic of each word in a document is not affected by the topic distribution of the other documents in BBM. It therefore avoids the problem of context ambiguity that BTM encounters when aggregating all the documents in a corpus. Moreover, it guarantees the model cohesion between the training and test phrases. 
\section{BBM in topic models}\label{sec:IV}
As we discussed, BBM can be applied to a large class of probabilistic models. In what follows, we show how to deploy BBM for the two conventional topic models: Latent Dirichlet Allocation (LDA) and Hierarchical Dirichlet Process (HDP).  
\subsection{Case study: Latent Dirichlet Allocation}\label{sec:bbm-lda}
\par
\begin{figure}[h]
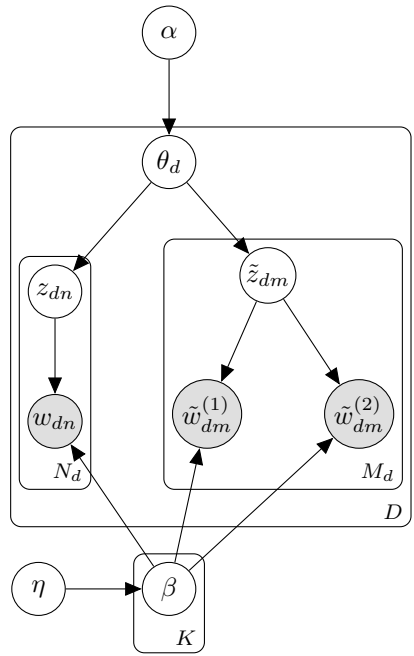

      \centering
      \tikz{ %
      	\node[latent] (eta) {$\eta$}; %
      	\node[latent, right= of eta] (beta) {$\beta$}; %
      	\node[obs, above= 1.5cm of beta, xshift=-1.5cm] (wN) {$w_{dn}$}; %
      	\node[latent, above=of wN] (zN) {$z_{dn}$}; %
      	\node[obs, above=1.5cm of beta, xshift=0.5cm] (bM1) {$\tilde{w}_{dm}^{(1)}$}; %
      	\node[obs, above=1.5cm of beta, xshift=2.5cm] (bM2) {$\tilde{w}_{dm}^{(2)}$}; %
      	\node[latent, above=of bM1, xshift=0.8cm] (zM) {$\tilde{z}_{dm}$}; %
      	\node[latent, above=of zN, xshift=1.5cm] (theta) {$\theta_{d}$}; %
      	\node[latent, above=of theta] (alpha) {$\alpha$}; %
      	\edge {eta} {beta}; %
      	\edge {beta} {wN,bM1,bM2}; %
      	\edge {zN} {wN}; %
      	\edge {zM} {bM1, bM2}; %
      	\edge {theta} {zN, zM}; %
      	\edge {alpha} {theta}; %
      	\plate {uni}{(zN)(wN)}{$N_d$}; %
      	\plate {bi}{(zM)(bM1)(bM2)}{$M_d$}; %
      	\plate {}{(uni)(bi)(theta)}{$D$}; %
      	\plate {}{(beta)}{$K$};
     }
     \caption{The graphical representation of LDA-B}
     \label{fig:BBM_LDA}
\end{figure}

We start by introducing a case study/implementation of BBM following the scheme of the LDA model, called LDA-B . The key idea is that we explicitly make a distinction between words and biterms during learning. The generative process is comparable to LDA, except that we generate $ N_{d} $ words and $ M_{d} $ biterms in document $d$.

The notations in the graphical models and the parameter inference equations are as follow. $D$ is the number of documents while $K$ is the number of topics. For each document $d$, the total number of words and biterms are $N_d$ and $M_d$ respectively. $\theta $ is the topic-document distribution, $\beta$ is the word-topic distribution, $z_{dn} $ is the topic assignment for the word $w_{dn}$, and $ \tilde{z}_{dm} $ is the topic assignment for the biterm $ \tilde{w}_{dm} $. We use a tilde symbol to denote the biterm version of that variable. Finally, $\tilde{w}_{dm}^{(1)}$ and $\tilde{w}_{dm}^{(2)}$ are the first and second word of the biterm $\tilde{w}_{dm}$ respectively.

\par
Figure \ref{fig:BBM_LDA} show the graphical representation of LDA-B. The generative process for LDA-B is as follow:
\begin{enumerate}
\item Draw topics $ \beta_{k} \sim \text{Dirichlet}(\eta) $ for $ k \in [1, K] $
\item For each document $ d \in [1,D] $:
	\begin{enumerate}
		\item Draw topic proportions $ \theta_{d} \sim \text{Dirichlet}(\alpha) $
		\item For each word $w_{dn} \in \{1,...,N_d\}$: 
		\begin{enumerate}
			\item Draw topic assignment $z_{dn} \sim \text{Multinomial}(\theta_d) $
			\item Draw word $w_{dn} \sim \text{Multinomial}(\beta_{z_{dn}})$
		\end{enumerate}
		\item For each biterm $\tilde{w}_{dm} \in \{1,...,M_d\}$:
		\begin{enumerate}
			\item Draw topic assignment $\tilde{z}_{dm} \sim \text{Multinomial}(\theta_d) $
			\item Draw two words $\tilde{w}_{dm}^{(1)}, \tilde{w}_{dm}^{(2)} \sim \text{Multinomial}(\beta_{\tilde{z}_{dm}})$
		\end{enumerate}
	\end{enumerate}
\end{enumerate}

\par
Based on the procedure above, our goal is to compute the posterior distribution. However, as this computation is intractable, we need to approximate it by a sampling or inference algorithms. In this work, we follow the mechanism of the primitive LDA \cite{blei2003latent} to perform the inference for both the term part and the biterm part of LDA-B. In particular, we choose the mean-field variational inference algorithm for this purpose. The details to compute the variational inference is shown in Appendix \ref{appenB}. The approximating distribution $q_D$ takes the following form:
\begin{align*}
q_{D} (\beta, \theta, z \mid \lambda, \gamma, \phi) & = \prod_{k=1}^{K} q_{D}(\beta_k \mid \lambda_k). \prod_{d=1}^{D} p(\theta_d \mid \gamma_d) \\ & . \prod_{d=1}^{D} \prod_{n=1}^{N_d} q_{D}(z_{dn} \mid \phi_{d, w_{dn}}). \prod_{d=1}^{D}  \prod_{m=1}^{M_d}q_{D} (\tilde{z}_{dm} \mid \tilde{\phi}_{d, \tilde{w}_{dm}})
\end{align*}
where $q_{D}(\beta_k \mid \lambda_k) = \text{Dirichlet}_{V} (\beta_k \mid \lambda_k)$, $q_D (\theta_d \mid \gamma_d) = \text{Dirichlet}_{K} (\theta_d \mid \gamma_d)$, and $q_{D}(z_{dn} \mid \phi_{dn}) = \text{Multinomial}_{K} (z_{dn} \mid \phi_{dn})$, $q_{D} (\tilde{z}_{dm} \mid \tilde{\phi}_{dm}) = \text{Multinomial}_{K} (\tilde{z}_{dm} \mid \tilde{\phi}_{dm})$. The subscripts on the Dirichlet and Multinomial denote the dimensions of distributions. Here we summarize the updating rules for the following variational parameters: $\lambda $ indicates the topic distribution over words, $ \gamma$ is the topic proportion in each document, $\phi$ and $\tilde{\phi} $ describe the assignment of each word and each biterm (in each document) to a topic respectively. The approximation of the lower bound for $q_D$ is described in Appendix \ref{appenB}. 

First, the update for $\gamma$ is:
\begin{equation}\label{eq:LDA-B-gamma}
\gamma_{d, k} = \alpha_{k} + \sum_{n=1}^{N_d} \phi_{d,n,k} + \sum_{m=1}^{M_d} \tilde{\phi}_{d,m,k}
\end{equation}

Then, the update for $\lambda$ is:
\begin{equation}\label{eq:LDA-B-lambda}
 \lambda_{k, v} \leftarrow  \eta_{k, v} + \sum_{d=1}^D f(v, \phi, \tilde{\phi})
\end{equation}
where, 
\begin{equation}
f(v, \phi, \tilde{\phi}) =  \sum_{n=1}^{N_d} I\{ w_{dn} = v\} \phi_{d,n,k} +  \sum_{m=1}^{M_d} \left[ I\{ \tilde{w}^{(1)}_{dm} = v \} + I\{ \tilde{w}^{(2)}_{dm} = v \} \right] \tilde{\phi}_{d,m,k}
\end{equation}

Next, we write the update for $\phi$:
\begin{equation}\label{eq:LDA-B-phi}
\phi_{d, n, k} \propto exp \lbrace E_{q}[\log \theta_{d, k}] + E_{q}[\log \beta_{k,w_{dn}}] \rbrace
\end{equation}

Finally, the update for $\tilde{\phi} $ can be written by:
\begin{equation}\label{eq:LDA-B-phi-2}
\tilde{\phi}_{d, m, k}  \propto exp \lbrace E_{q}[\log \theta_{d, k}] + E_{q}[\log \beta_{k,w_{dm}^{(1)}}] + E_{q}[\log \beta_{k,w_{dm}^{(2)}}] \rbrace
\end{equation}

\begin{figure}[ht]
\begin{minipage}[t]{0.45\linewidth}
\begin{algorithm}[H]
\caption{$\text{LocalVB}(d, \lambda)$}
\label{algo:LOCAL_VB}
\begin{algorithmic}
\State Initialize : $\gamma_d$
	\While {$(\gamma_d, \phi_d, \tilde{\phi}_d)$ not converged}
	\State {$\forall (k,n)$, set $\phi_{dnk}$ from \eqref{eq:LDA-B-phi}}
	\State {$\forall (k,m)$, set $\tilde{\phi}_{dmk}$ from \eqref{eq:LDA-B-phi-2}}
	\State {(normalized across k)}
	\State {$\forall k$, set $\gamma_{dk}$ from \eqref{eq:LDA-B-gamma}}
	\EndWhile
	\State \Return $(\gamma_d, \phi_d, \tilde{\phi}_d)$
\end{algorithmic}
 \end{algorithm}
\end{minipage}
\hspace{0.2cm}
\begin{minipage}[t]{0.55\linewidth}
\begin{algorithm}[H]
\caption{SVI for LDA-B}
\label{algo:SVI_LDA}
\begin{algorithmic}
\Require{ Hyperparameters $\eta, \alpha, D, (\rho)_{t=1}^T$}
\Ensure{$\lambda$}
\State Initialize : $\lambda$
\For{$t=1,...,T$}
\State {Collect new data minibatch C}
\State {$\rho_t = (\tau + t)^{-\kappa} $}
\ForEach {document indexed d in C}
\State {$(\gamma_d, \phi_d, \tilde{\phi}_d) \leftarrow \text{LocalVB} (d, \lambda)$}
\EndFor
\State {$\forall (k,v)$, $\tilde{\lambda}_{kv} \leftarrow \eta_{kv} + \frac{D}{|C|} \sum_{\text{d in C}} f(v, \phi, \tilde{\phi})$}
\State {$\forall (k,v), \lambda_{kv} \leftarrow (1 - \rho_t) \lambda_{kv} + \rho_t \tilde{\lambda}_{kv}$}
\EndFor
\end{algorithmic}
 \end{algorithm}
\end{minipage}

\begin{minipage}[b]{0.45\linewidth}
\begin{algorithm}[H]
\caption{SVB for LDA-B}
\label{algo:SVB_LDA}
\begin{algorithmic}
\Require{ Hyperparameters $\eta, \alpha$}
\Ensure{A sequence $\lambda^{(1)}, \lambda^{(2)},...$}
\State {Initialize $\forall (k,n), \lambda_{kn}^{(0)} \leftarrow \eta_{kn}$}
\For{$b=1,2,...$}
\State {Collect new data minibatch C}
\ForEach {document d in C}
\State {$(\gamma_d, \phi_d, \tilde{\phi}_d) \leftarrow \text{LocalVB} (d, \lambda)$}
\EndFor
\State {$\forall (k,v)$, $\tilde{\lambda}_{kv} \leftarrow \sum_{\text{d in C}} f(v, \phi, \tilde{\phi})$}
\State {$\forall (k,v), \lambda_{kv}^{(b)} \leftarrow \lambda_{kv}^{(b-1)} + \tilde{\lambda}_{kv}$}
\EndFor
\end{algorithmic}
 \end{algorithm}
\end{minipage}
\hspace{0.2cm}
\begin{minipage}[b]{0.55\linewidth}
\begin{algorithm}[H]
\caption{KPS for LDA-B}
\label{algo:KPS_LDA}
\begin{algorithmic}
\Require{ Hyperparameters $\eta, \alpha$}
\Ensure{$\lambda$}
\State Initialize : $\lambda_0 \leftarrow \eta$
\For{$b=1,2,...$}
\State {Collect new data minibatch C}
\ForEach {document d in C}
\State {$(\gamma_d, \phi_d, \tilde{\phi}_d) \leftarrow \text{LocalVB} (d, \lambda)$}
\EndFor
\State {$\forall (k,v)$, $\tilde{\lambda}_{kv} \leftarrow \sum_{\text{d in C}} f(v, \phi, \tilde{\phi})$}
\State {$\forall (k,v), \lambda_{kv}^{(b)} \leftarrow \lambda_{kv}^{(b-1)} + \tilde{\lambda}_{kv} + \eta$}
\EndFor
\end{algorithmic}
 \end{algorithm}
\end{minipage}
\label{fig:alg-BBM}
\end{figure}

Based on the inference of LDA-B, we can extend the previous works in \cite{hoffman2010online,broderick2013streaming,duc2017keeping} to fit LDA-B in the online and streaming environment. The resulting algorithms is presented in Algorithm \ref{algo:LOCAL_VB}, \ref{algo:SVI_LDA}, \ref{algo:SVB_LDA}, \ref{algo:KPS_LDA}. Algorithm \ref{algo:LOCAL_VB} shows a variational Bayes procedure which takes the global parameter $\lambda$ to infer the topic distributions of the terms ($\phi$) and the biterms ($\tilde{\phi}$) of the document $d$. \textbf{Stochastic variational inference (SVI)} \cite{hoffman2010online} is an algorithm which enables LDA for online environment. We develop SVI to fit the LDA-B model as shown in Algorithm \ref{algo:SVI_LDA}. A worthy note is that SVI uses two parameters, $\tau  $ and $\kappa$, to determine the learning rate $\rho_t$ at the iteration $t$: $\rho_t = (\tau + t)^{-\kappa}$. Algorithm \ref{algo:SVB_LDA} presents the streaming algorithm (also called SVB-B) created by exploiting \textbf{Streaming Variational Bayes (SVB)} \cite{broderick2013streaming} as a base framework fitting LDA-B to the streaming environment. SVB-B is straightforward LDA-B except that SVB-B updates the local parameters for each minibatch of documents as they arrive and then updates the current estimate of $\lambda$ for each document $d$ in the minibatch. Finally, Algorithm \ref{algo:KPS_LDA} introduces the streaming algorithm (called KPS-B) developed from the \textbf{Keeping Priors in Streaming (KPS)} \cite{duc2017keeping} framework and the model LDA-B. KPS-B is similar to SVB-B except that it exploits human knowledge as a prior and keep it for each update. Therefore, the only distinction between Algorithm \ref{algo:KPS_LDA} and Algorithm \ref{algo:SVB_LDA} is that KPS-B keeps the impact of the prior $\eta$ on each minibatch instead of forgetting it as in SVB-B. 
\subsection{Case study: Hierarchical Dirichlet Process}
\begin{figure}[h]
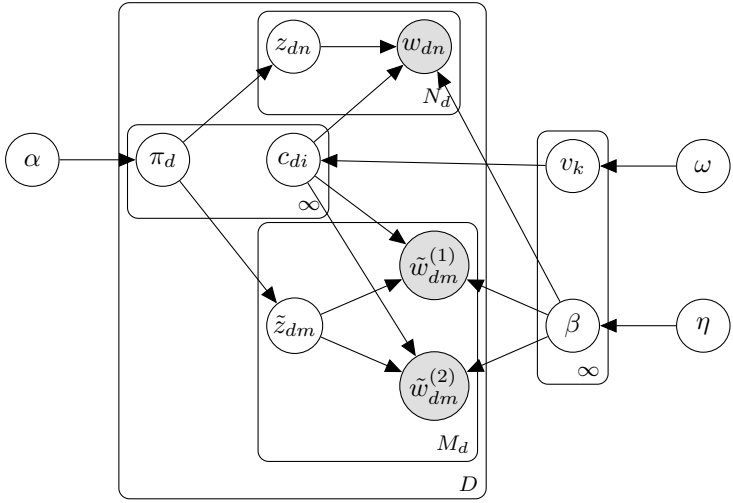

      \centering
      \tikz{ %
      	\node[latent] (alpha) {$\alpha$}; %
      	\node[latent, right = of alpha] (pi) {$\pi_d$}; %
      	\node[latent, right = of pi, yshift = 1.5cm] (zDN) {$z_{dn}$};
      	\node[latent, right = of pi, yshift = -2.2cm] (zDM) {$\tilde{z}_{dm}$};
      	\node[obs, right = of zDN] (w0) {$w_{dn}$};
      	\node[obs, right = of zDM, yshift = 0.8cm] (w1) {$\tilde{w}_{dm}^{(1)}$};
      	\node[obs, right = of zDM, yshift = -0.8cm] (w2) {$\tilde{w}_{dm}^{(2)}$};
      	\node[latent, right = of pi] (cDI) {$c_{di}$}; %
      	\node[latent, right = of w1, yshift = -0.8cm] (beta) {$\beta$};
      	\node[latent, right = of beta] (eta) {$\eta$};
      	\node[latent, above = of beta, yshift = 0.4cm] (vK) {$v_k$}; %
      	\node[latent, right = of vK] (omega) {$\omega$};
      	\edge {eta} {beta}; %
      	\edge {vK} {cDI}; %
      	\edge {omega} {vK};
      	\edge {beta, cDI} {w0, w1, w2};
      	\edge {zDN} {w0};
      	\edge {zDM} {w1, w2};
      	\edge {pi} {zDN, zDM};
      	\edge {alpha} {pi};
      	\plate {uni}{(zDN)(w0)}{$N_d$}; %
      	\plate {bi}{(zDM)(w1)(w2)}{$M_d$}; %
      	\plate {inf1}{(pi)(cDI)}{$\infty$};
      	\plate {}{(beta)(vK)}{$\infty$};
      	\plate {}{(uni)(bi)(inf1)}{$D$}; %
      	
     }
     \caption{The graphical representation of HDP-B}
     \label{fig:BBM_HDP}
\end{figure}
Similar to LDA-B, this section describes HDP-B, a case study/implementation of BBM following the scheme of the HDP-B model. Figure \ref{fig:BBM_HDP} shows the graphical representation of HDP-B. The generative process of HDP-B is as follow:
\begin{enumerate}
\item Draw an infinite number of topics, $ \beta_k \sim \text{Dirichlet}(\eta) $ for $ k \in \lbrace 1,2,3,...\rbrace $
\item Draw corpus breaking proportions, $ v_k \sim \text{Beta}(1, \omega) $ for  $ k \in \lbrace 1,2,3,...\rbrace $
\item For each document $d$:
		\begin{enumerate}
			\item Draw document-level topic indices, $c_{d,i} \sim \text{Multinomial}(\sigma(v)) $ for  $ i \in \lbrace 1,2,3,...\rbrace $
			\item Draw document breaking proportions, $ \pi_{d,i} \sim \text{Beta}(1, \alpha) $ for  $ i \in \lbrace 1,2,3,...\rbrace $
			\item For each word $w_{dn}$:
			\begin{enumerate}
				\item Draw topic assignment $ z_{dn} \sim \text{Multinomial}(\sigma(\pi_d)) $
				\item Draw word $w_{dn} \sim \text{Multinomial}(\beta_{c_{d, z_{dn}}})$ 
			\end{enumerate}

			\item For each biterm $\tilde{w}_{dm}$:
			\begin{enumerate}
				\item Draw topic assignment $ \tilde{z}_{dm} \sim \text{Multinomial}(\sigma(\pi_d)) $
				\item Draw two words $ \tilde{w}_{dm}^{(1)}, \tilde{w}_{dm}^{(2)} \sim \text{Multinomial}(\beta_{c_{d, \tilde{z}_{dm}}})$ 
			\end{enumerate}
		\end{enumerate}
\end{enumerate}

Following the procedure for LDA-B, we apply mean-field variational inference to approximate the posterior distribution for HDP-B. The approximation for the biterm part of HDP-B is very similar to that of LDA-B. Likewise, the approximation for the word part is following the approximation in HDP (\cite{hoffman2013stochastic}) directly. We show the online learning algorithm for HDP-B in Algorithm \ref{algo:SVI_HDP}. The relevant information of parameters and expectations for each update are found in \cite{hoffman2013stochastic}.

\begin{algorithm}[H]
\caption{SVI for HDP-B}
\label{algo:SVI_HDP}
\begin{algorithmic}[1]
\State Initialize : $\lambda^{(0)}$ randomly. Set $a^{(0)}=1$ and $b^{(0)}= \omega$
\State Set the step-size schedule $\rho_t$ appropriately 
\Repeat
\State Sample a document $d$ uniformly from the data set
\State For $i \in \{1,...,T\}$ initialize 
$$ \zeta_{di}^{k} \propto \exp \left\lbrace \sum_{n=1}^{N_d} E[\log \beta_{k,w_{dn}}] + \sum_{m=1}^{M_d} \left( E[\log \beta_{k,\tilde{w}_{dm}^{(1)}}] + E[\log \beta_{k,\tilde{w}_{dm}^{(2)}}] \right) \right\rbrace , k \in \{1,...,K\} $$
\State For $n \in \{1,2,...,N_d\}$ initialize
$$\textstyle \phi_{dn}^{i} \propto \exp \left\lbrace \sum_{k=1}^{K} \zeta_{di}^{k} E[\log \beta_{k,w_{dn}}] \right\rbrace, i \in \{1,...,T\} $$
\State For $m \in \{1,2,...,M_d\}$ initialize
$$\textstyle \tilde{\phi}_{dm}^{i} \propto \exp \left\lbrace \sum_{k=1}^{K} \zeta_{di}^{k} \left( E [\log \beta_{k,\tilde{w}_{dm}^{(1)}}] + E [\log \beta_{k,\tilde{w}_{dm}^{(2)}}] \right) \right\rbrace, i \in \{1,...,T\} $$

\algstore{myalg}
\end{algorithmic}
\end{algorithm}

\begin{algorithm}[H]                  
\begin{algorithmic} [1]                   
\algrestore{myalg}

\Repeat
\State For $i \in \{1,...,T\}$ set
\begin{align*}
& \textstyle \gamma_{di}^{(1)} = 1 + \sum_{n=1}^{N_d} \phi_{dn}^{i} + \sum_{m=1}^{M_d} \tilde{\phi}_{dm}^{i} \\ & \textstyle \gamma_{di}^{(2)} = \alpha + \sum_{n=1}^{N} \sum_{j=i+1}^{T} \phi_{dn}^{j} + \sum_{m=1}^{M_d} \sum_{j=i+1}^{T} \tilde{\phi}_{dm}^{j} \\ & \zeta_{di}^{k} \propto \exp \left\lbrace E[\log \sigma_k (V)] + \sum_{n=1}^{N_d} \phi_{dn}^{i} E[\log \beta_{k,w_{dn}}] + \sum_{m=1}^{M_d} \tilde{\phi}_{dn}^{i} \left( E[\log \beta_{k,\tilde{w}_{dm}^{(1)}}] + E[\log \beta_{k,\tilde{w}_{dm}^{(2)}}] \right) \right\rbrace \\ & \text{for k} \in \{1,...,K\} 
\end{align*}
\State For $n \in \{1,2,...,N_d\}$ set
$$\textstyle \phi_{dn}^{i} \propto \exp \left\lbrace E[\log \sigma_i (\pi_d)] +  \sum_{k=1}^{K} \zeta_{di}^{k} E[\log \beta_{k,w_{dn}}] \right\rbrace, i \in \{1,...,T\} $$
\State For $m \in \{1,2,...,M_d\}$ set
$$\textstyle \tilde{\phi}_{dm}^{i} \propto \exp \left\lbrace E[\log \sigma_i (\pi_d)] + \sum_{k=1}^{K} \zeta_{di}^{k} \left( E [\log \beta_{k,\tilde{w}_{dm}^{(1)}}] + E [\log \beta_{k,\tilde{w}_{dm}^{(2)}}] \right) \right\rbrace, i \in \{1,...,T\} $$
\Until local parameters converge
\State For $k \in \{1,...,K\}$ set intermediate topics
\begin{align*}
 \textstyle \hat{\lambda}_{kv} & = \eta + D \sum_{i=1}^{T} \zeta_{di}^{k} \left( \sum_{n=1}^{N_d} \phi_{dn}^i I[w_{dn}=v] + \sum_{m=1}^{M_d} \tilde{\phi}_{dm}^{i} (I[\tilde{w}_{dm}^{(1)}=v] + I[\tilde{w}_{dm}^{(2)}=v])\right) \\  \textstyle \hat{a}_k &= 1 + D \sum_{i=1}^{T} \zeta_{di}^{k} \\  \textstyle \hat{b}_k &= \omega + D \sum_{i=1}^{T} \sum_{l=k+1}^{K} \zeta_{di}^{l}
\end{align*}
\State Set
\begin{align*}
& \textstyle \lambda^{(t)} = (1 - \rho_t) \lambda^{(t-1)} + \rho_t \hat{\lambda} \\ & \textstyle \hat{a} = (1 - \rho_t) a^{(t-1)} + \rho_t \hat{a} \\ &  \textstyle \hat{b} = (1 - \rho_t) b^{(t-1)} + \rho_t \hat{b}
\end{align*}
\Until forever
\end{algorithmic}
\end{algorithm}

\section{Experimental evaluation}
\label{sec:V}
This section evaluates the effectiveness of the BoB representation and the BBM framework with different premitive topic models on four large scale datasets. In particular, we compare the performance of the representations BoB and BoW in two tasks: the unsupervised topic modeling with LDA \cite{blei2003latent} and HDP \cite{teh2006hierarchical}; and text classification with Support Vector Machines (SVM) \cite{CC01a}. In addition, we investigate the performance of BoB for the normal texts to see if BoB can be a general method for different kind of texts (i.e, both short and and usual long texts). Regarding BBM, we compare it with the base topic models to demonstrate its effectiveness for short texts. First, we compare HDP-B with the primitive HDP using both the BoB and BoW representations. Second, we evaluate the performance of LDA-B in the context of online and streaming environment.
\par
\textbf{Datasets:} We use $4$ large scale short-text datasets in this work. Apart from the standard short-text dataset  \textit{TMNtitle}\footnote{\url{http://acube.di.unipi.it/TMNdataset}}, we employ three other large collections of short texts that have been crawled on the Web. These three datasets are: \textit{Yahoo Questions}\footnote{\url{https://answers.yahoo.com/}}, \textit{Tweets}\footnote{\url{http://twitter.com/}} and \textit{Nytimes Titles}\footnote{\url{http://www.nytimes.com/}}. The details about the crawling process are described in \cite{mai2016enabling}. We summarize several important points of the four datasets in Table \ref{tab:dataset_description}. These datasets went through a preprocessing pipeline that includes tokenizing, stemming, removing stopwords, removing low-frequency words (i.e, appearing in less than 3 documents), and removing extremely short documents (less than 3 words).
\begin{table}
\centering 
\caption{Summary of 4 text datasets, V is the size of vocabulary}
\label{tab:dataset_description}
\centering
\begin{tabular}{|c|c|c|c|c|c|}
\hline
& Corpus size & Average length per doc & V & Number of labels \\ \hline
Yahoo Questions &537,770&4.73&24,420&20 \\ \hline
Tweets &1,485,068&10.14&89,474&69 \\ \hline
Nytimes Titles & 1,684,127&5.15&55,488&14 \\ \hline 
TMNtitle & 26,251 & 4.6 & 2,823 & 7 \\ \hline
\end{tabular}
\end{table}
\par
\textbf{Performance measure:} The common performance measure to evaluate topic models is Log Predictive Probability (LPP) that measures the predictiveness and generalization of a learned model to new data. The procedure to compute this measure is introduced in \cite{hoffman2013stochastic}. For each dataset, we randomly divide it into two parts $D_{train}$ and $D_{test}$. We only hold documents whose length is greater than 4 and randomly divide it into two parts ($\mathbf{w}_1,\mathbf{w}_2$) with ratio $4:1$ (4 : 1 is chosen to make sure that $\mathbf{w}_2$ has at least 1 term). We do inference for $\mathbf{w}_1$ and estimate the probability of $\mathbf{w}_2$ given $\mathbf{w}_1$. In the case of BoB, we first need to convert the topic-over-biterms (distributions over biterms) to topic-over-words (distributions over words) whose conversion formula is described in Appendix \ref{appenA}. Based on these notations, LPP is computed as follows:
\begin{equation}
\text{LPP}(\mathcal{D}_{test}) = \frac{\sum_{d \in \mathcal{D}_{test}}\text{LPP}(d)}{|\mathcal{D}_{test}|}
\end{equation}
where $|\mathcal{D}_{test}|$ is the number of documents in $\mathcal{D}_{test}$, and $\text{LPP}(d)$ for each document $d$ in the test set is computed by:
\begin{equation}
\text{LPP}(d) = \frac{\sum_{w \in \mathbf{w}_2}\log p(w \mid \mathbf{w}_1, \mathcal{D}_{train})}{|\mathbf{w}_2|}
\end{equation}
Here, $|\mathbf{w}_2|$ is the number of words in $\mathbf{w}_2$ and:
\begin{align*}
p(w \mid \mathbf{w}_1, \mathcal{D}_{train}) &\approx p(w\mid \hat{\boldsymbol\pi})\\
&= \sum_{k}^{K}p(w \mid z = k)p(z=k \mid \hat{\boldsymbol\pi}) = \sum_{k}^{K}\phi_{kw}\hat{\pi}_k\\
\end{align*}

\subsection{Evaluation of BoB representation}
In this section, we compare BoB to BoW in two tasks: topic modeling and classification. In each task, we choose different biterm thresholds to make sure that the number of generated biterms is not too large. A biterm threshold \cite{mai2016enabling} is defined as the least value of the number of documents containing a specific biterm.  Consequently, a biterm would be removed if the number of documents containing it is less than the biterm threshold.

\subsubsection{BoB in topic modeling}
\label{sub:bob-topicModel}
\textbf{Baseline Methods:} We use the BoW representation as the baseline to compare with BoB due to the widespread of the application of BoW in topic modeling. 
\par 
\textbf{Models in use:} In order to evaluate the performance of the new representation, we run the online HDP and online LDA\footnote{We use the source code of Online LDA and Online HDP from \url{https://github.com/Blei-Lab}} over each dataset with various settings for each representation (i.e, BoW and BoB). At time $t$, the fast noisy estimates of gradient is computed by subsampling a small set of documents (called minibatch). Based on such noisy estimates, the intermediate global parameters are calculated, followed by the last step to update the global parameters with a decreasing learning rate schedule $ \rho_t \leftarrow (\tau + t)^{-\kappa}$  ($\kappa$ is forgetting rate and $\tau$ is the delay).
\par
\textbf{Settings:} The parameters ($\tau,\kappa$) in both Online LDA and Online HDP form a grid: $\tau\in\{1,20, 40, 60, 80, 100\}$, $\kappa\in\{0.6, 0.7, 0.8, 0.9\}$. For each combination of $(\tau, \kappa)$, we fix the minibatch size of 5000 for three datasets \textit{Yahoo Questions}, \textit{Tweets} and \textit{Nytimes Titles}. For \textit{TMNtitle}, the batchsize is set to 500 due to the minor size of dataset. For online HDP, we set the truncation for corpus $K$=100, truncation for document $T$ = 20, $\eta =0.01$, $\alpha_0 = 1.0$, and $\omega = 1.0$. For online LDA, the number of topics $K$ is set to 100 for the $3$ datasets \textit{Yahoo}, \textit{Tweets}, \textit{Nytimes Titles} and $K=50$ for \textit{TMN title}. The biterm thresholds for different datasets are shown in Table \ref{biterm_table}.
\begin{table}[t]
\caption{Biterm threshold and vocabulary size in BoB ($V_b$)}
\label{biterm_table}
\resizebox{\textwidth}{!}{
\begin{tabular}{|c|c|c|c|c|}
\hline
& Yahoo Questions & Tweets & Nytimes Titles & TMNtitle \\ \hline
V(number of distinct words) & 24,420&89,474&55,488 & 2,823 \\ \hline
Biterm threshold & 2 & 10 & 5 & 2 \\ \hline

$V_b$ (number of distinct biterms) & 722,238 &764,385 &756,700 & 14,799 \\ \hline

\end{tabular}
}
\end{table}

\begin{figure*}
\begin{center}
\includegraphics[scale=0.23]{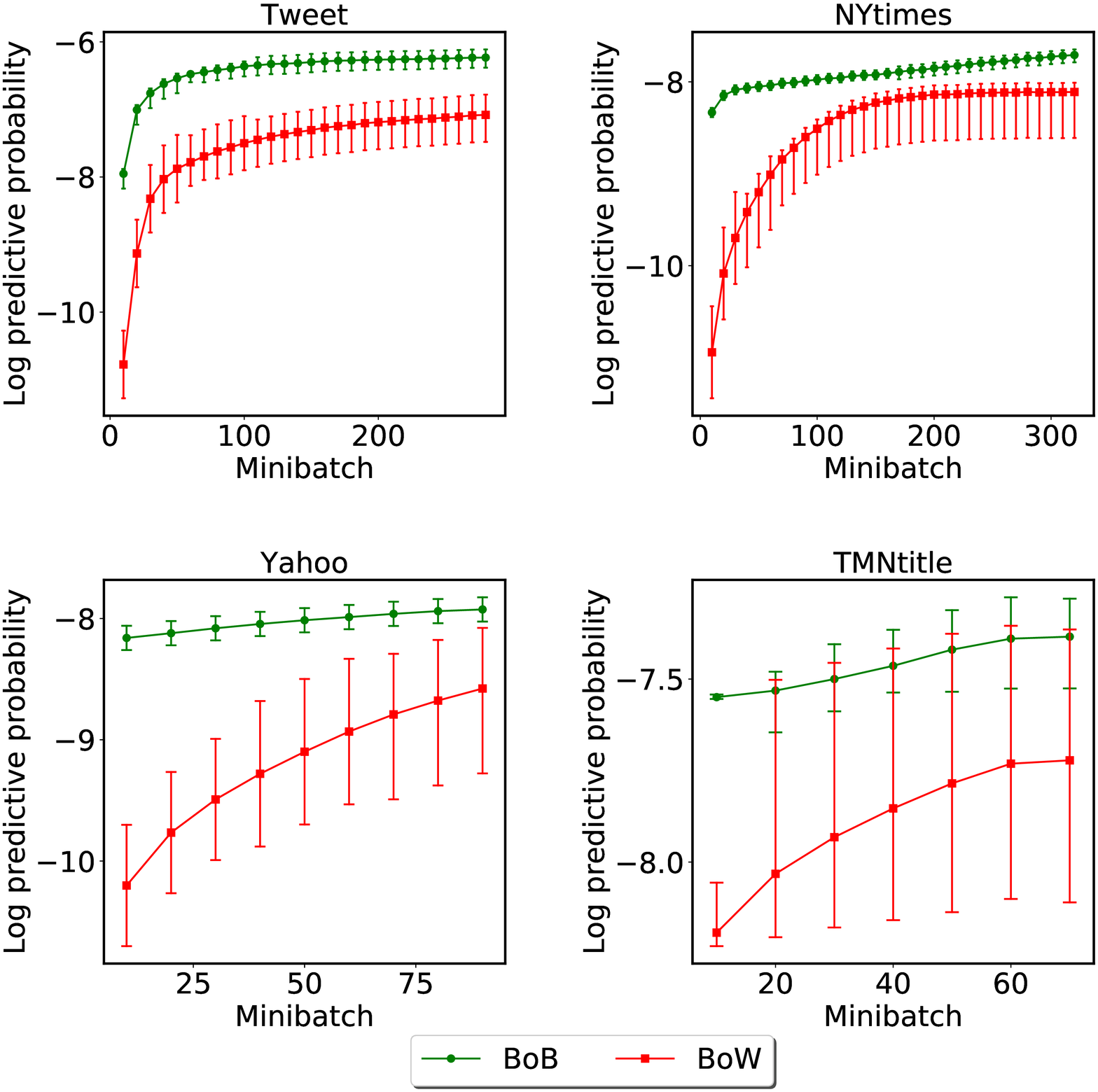}
\end{center}
\caption{The range of predictiveness (associated with the 24 parameter combinations) on four datasets for both BoW and BoB in online HDP. The lines are the averages of the 24 LPPs corresponding to the 24 combinations. The higher is the better.}
\label{fig:hdp_perplexity}
\end{figure*}

\begin{figure*}
\begin{center}
\includegraphics[scale=0.23]{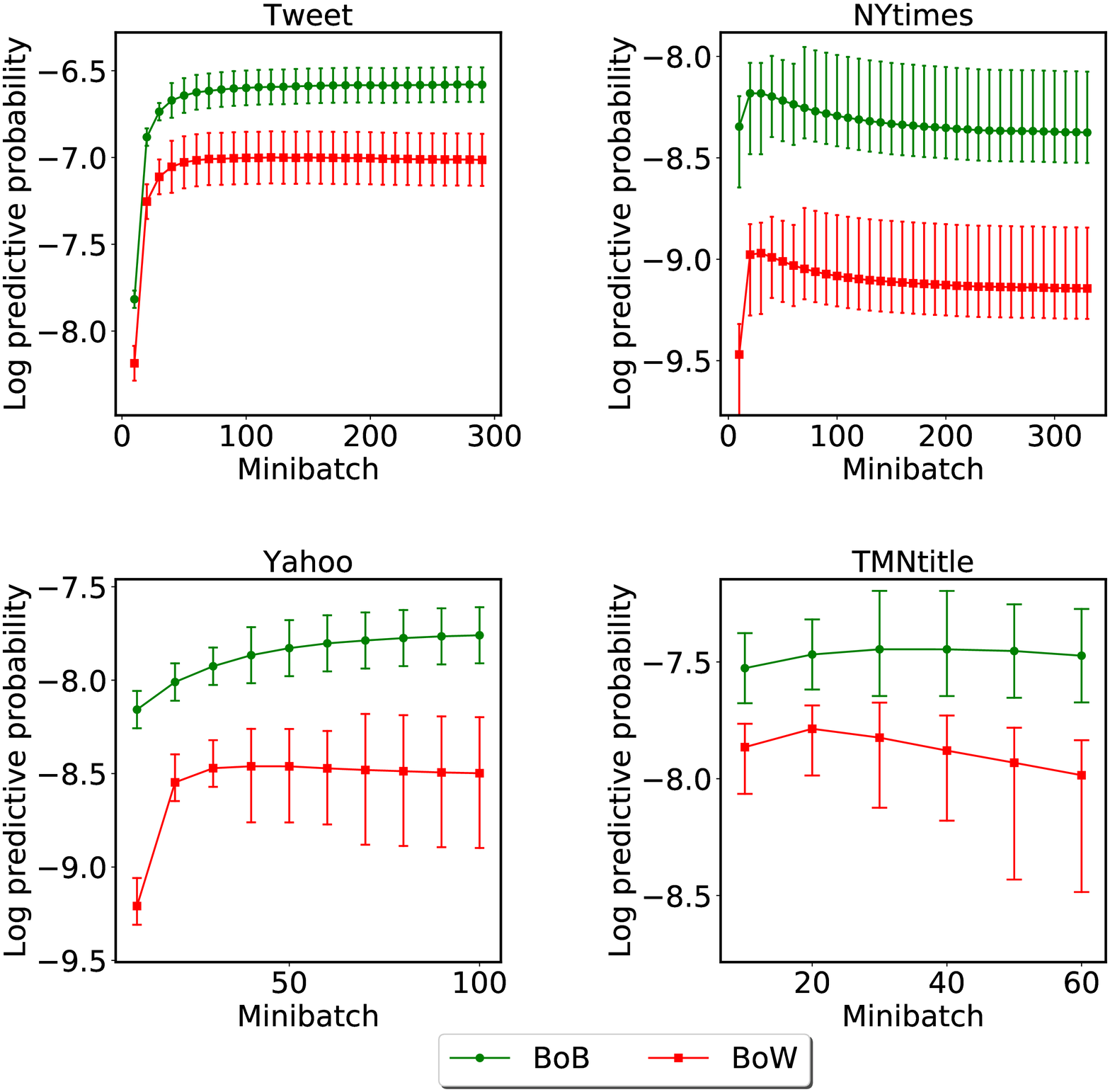}
\end{center}
\caption{The range of predictiveness (associated with 24 parameter combinations) on four datasets for both BoW and BoB in online LDA. The lines are the averages of the 24 LPPs corresponding to the 24 combinations. The higher is the better.}
\label{fig:lda_perplexity}
\end{figure*}
\par
\textbf{Experimental Results:} Figure \ref{fig:hdp_perplexity} and Figure \ref{fig:lda_perplexity} show the averages (the line) and the range of 24 LPPs associated with the 24 combinations of ($\tau, \kappa $) on four datasets (respectively).
\par 
Considering the online HDP model in Figure \ref{fig:hdp_perplexity}, we see that the predictive capability of online HDP with both BoB and BoW is improved in accordance with the increase of the number of learned documents for all the datasets. In every experimental scenario, the models with BoB always outperforms those with BoW significantly. In addition, we see that BoB always has better starting points than BoW across all the datasets and topic models. This implies that the models with BoW would require less training documents to converge than the models with BoW. These results overall suggest that BoB is more effective than BoW in modeling short texts with online HDP.

\par
For the online LDA in Figure \ref{fig:lda_perplexity}, the performance of BoB and BoW with the \textit{Tweets} and \textit{Yahoo Questions} datasets increase along with the number of documents being learned. Whereas, for the \textit{Nytimes Titles} and textit{TMNtitle}, the two figures for BoB and BoW reach a peak and then decrease slightly. However, the predictive capacity of BoB always outperforms BoW for all the 24 LPPs. In addition, BoB also has better starting points than BoW, which is similar to the result of Online HDP.

\subsubsection{Evaluating BoB in supervised methods: text classification with SVM}
In this part, we compare the performance of BoB and BoW in the text classification task using Support Vector Machine (SVM) \cite{CC01a}. We consider two types of weighting schemas for the terms/biterms in the documents: \textit{tf} (term frequency) and \textit{tf-idf} (term frequency-inverse document frequency). We also carry out the experiments that employ BoB for normal-text datasets to see the potential of BoB as a general representation for text analysis. For \textit{tf}, we normalize the frequency vectors by dividing the frequencies of word/term by the length of the document. 
We use the following formula to compute \textit{tf-idf} weight for BoW: 
\begin{equation}
\text{weight}(w,d) = tf(w,d)\log\left(\frac{N}{df(w)} \right)
\end{equation}
where $w$ is a word in the document $d$, $N$ is the corpus size, $tf(w,d)$ is the frequency of $w$ in $d$, and $df(w)$ is the number of documents containing $w$ in the corpus. 
\par 
For BoB, as described in Section \ref{sec:III}, the \textit{tf-idf} weight of biterm $(w_i, w_j)$ is computed by $\text{weight} (w_i, w_j) = \min(\text{weight} (w_i), \text{weight} (w_j))$.
\par
\textbf{Datasets: } For short-text datasets, we use the datasets \textit{Yahoo Questions} and \textit{Nytimes Titles} while the two popular datasets \textit{20Newsgroup} and \textit{Ohscal} are chosen for the case of normal texts. The detailed information about  these two normal-text datasets is shown in Table  \ref{tab:long_dataset_desc}.
 \begin{itemize}
\item \textit{20Newsgroup\footnote{\url{http://qwone.com/~jason/20Newsgroups/}}}- example of medium texts: collection of about 20000 messages taken from 20 Usenet newsgroups. 
\item \textit{ Ohscal\footnote{\url{http://glaros.dtc.umn.edu/gkhome/fetch/sw/cluto/datasets.tar.gz}}} - example of long texts: is subset of OHSUMED\footnote{\url{http://davis.wpi.edu/xmdv/datasets/ohsumed.html}}-collections of medical abstracts from MEDLINE\footnote{\url{https://www.medline.com}} database. 
\end{itemize}

For each dataset, we randomly divide it into five equal parts and report the accuracy of the models using 5-fold cross validation. For the SVM model, we employ the linear SVM implemented in the LIBLINEAR\footnote{\url{https://www.csie.ntu.edu.tw/~cjlin/libsvm/}} toolkit.

 \par 
\begin{table}
\centering
\caption{Description of two datasets used in regular text classification}
\label{tab:long_dataset_desc}
\resizebox{\textwidth}{!}{
\begin{tabular}{|c|c|c|c|c|}
\hline
Dataset & corpus size &  vocabulary size & Average length per doc& number of labels \\ \hline
Ohscal& 11,162& 11,465& 60.42 &10\\ \hline
20Newsgroup&19,928&62,061&79.97&20\\ \hline
\end{tabular}
}
\end{table}
\begin{figure*}[t]
\begin{center}
\subfloat[Yahoo Questions]{\includegraphics[scale=0.3]{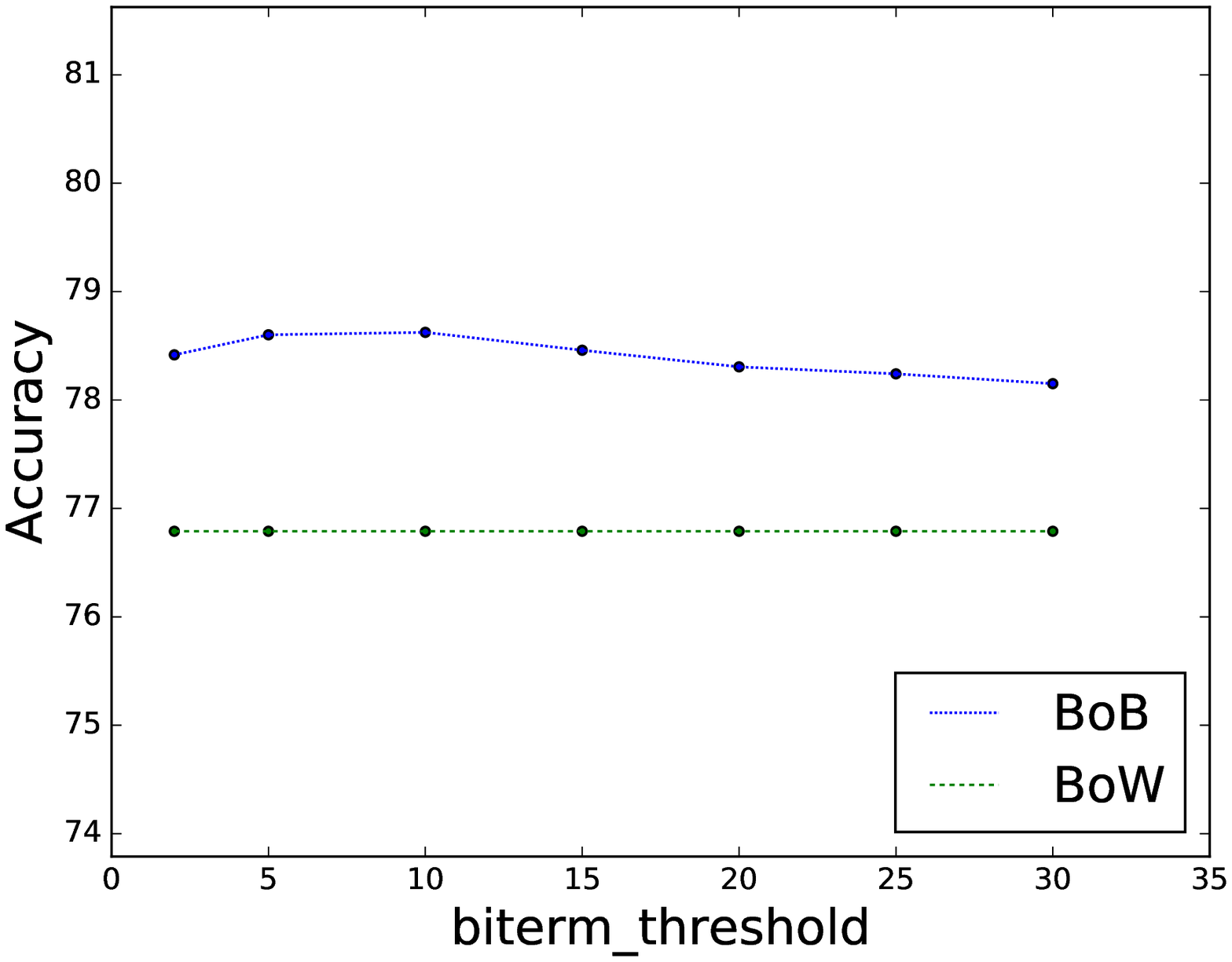}}
\subfloat[Nytimes Titles]{\includegraphics[scale=0.3]{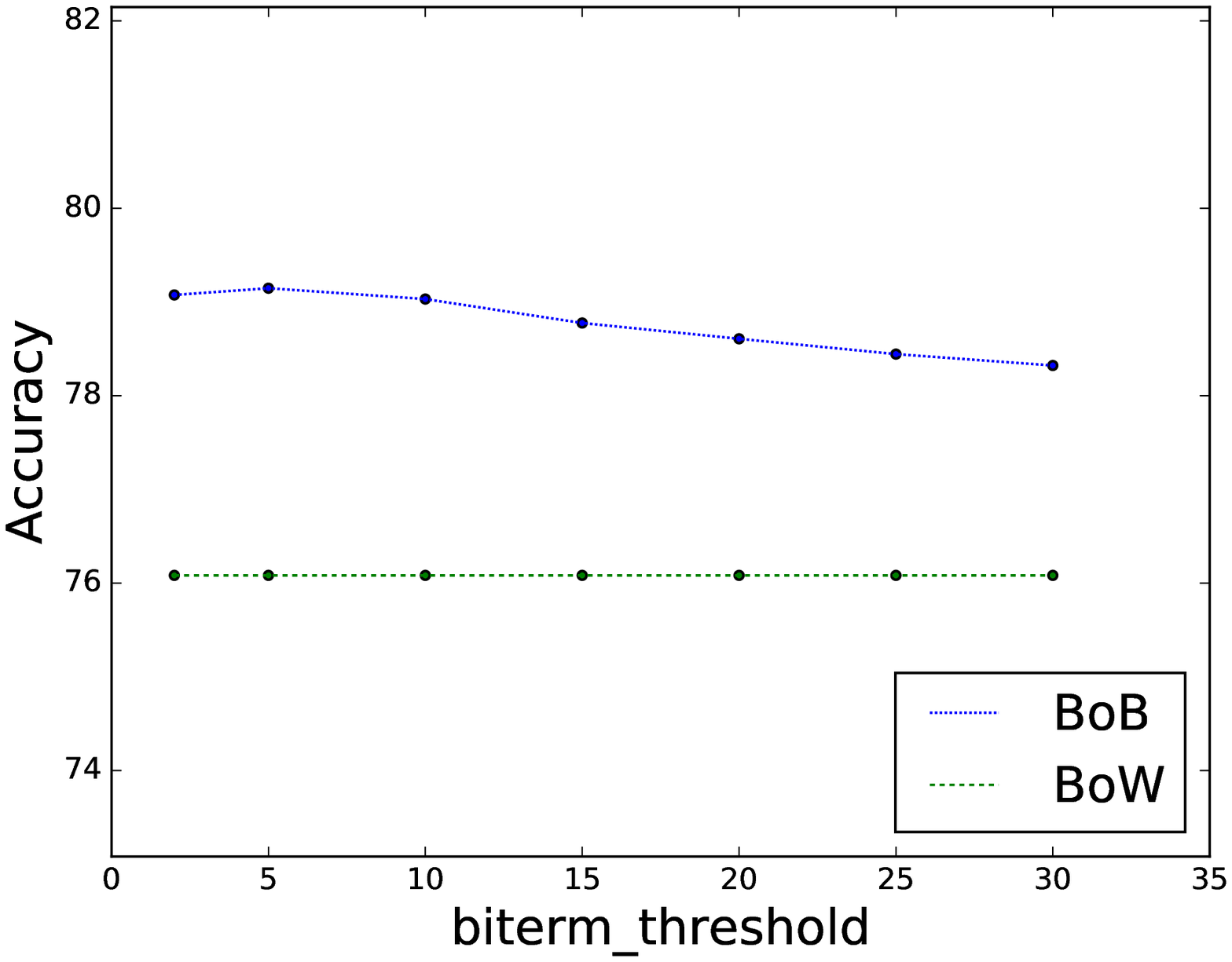}}
\end{center}
\caption{Classification performance on \textit{Yahoo Questions} and \textit{Nytimes Titles} with the weighting schema \textit{tf} and the document representations BoB and BoW.}
\label{fig:classification_tf}
\end{figure*}

\begin{figure*}
\begin{center}
\subfloat[Yahoo Questions]{\includegraphics[scale=0.3]{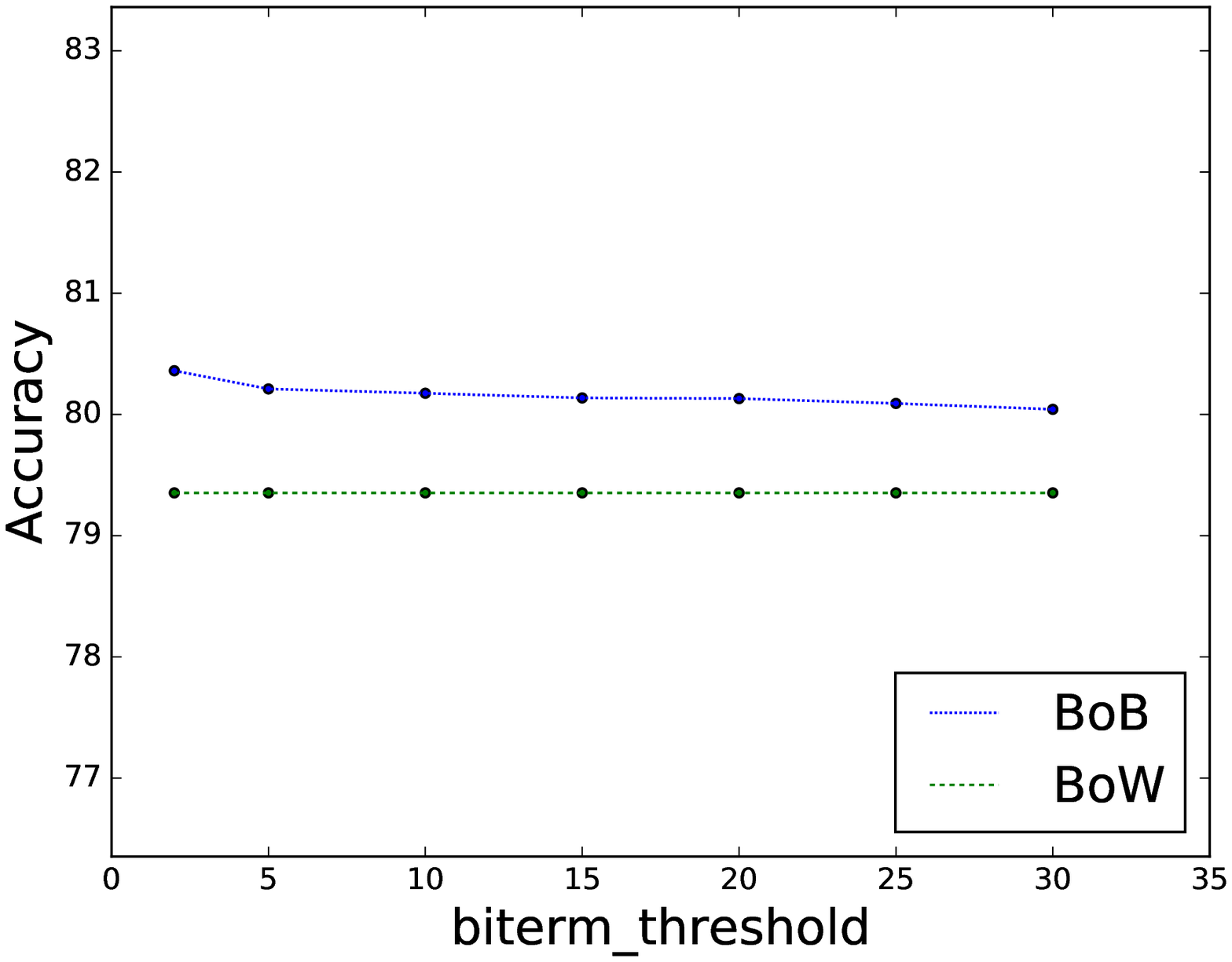}}
\subfloat[Nytimes Titles]{\includegraphics[scale=0.3]{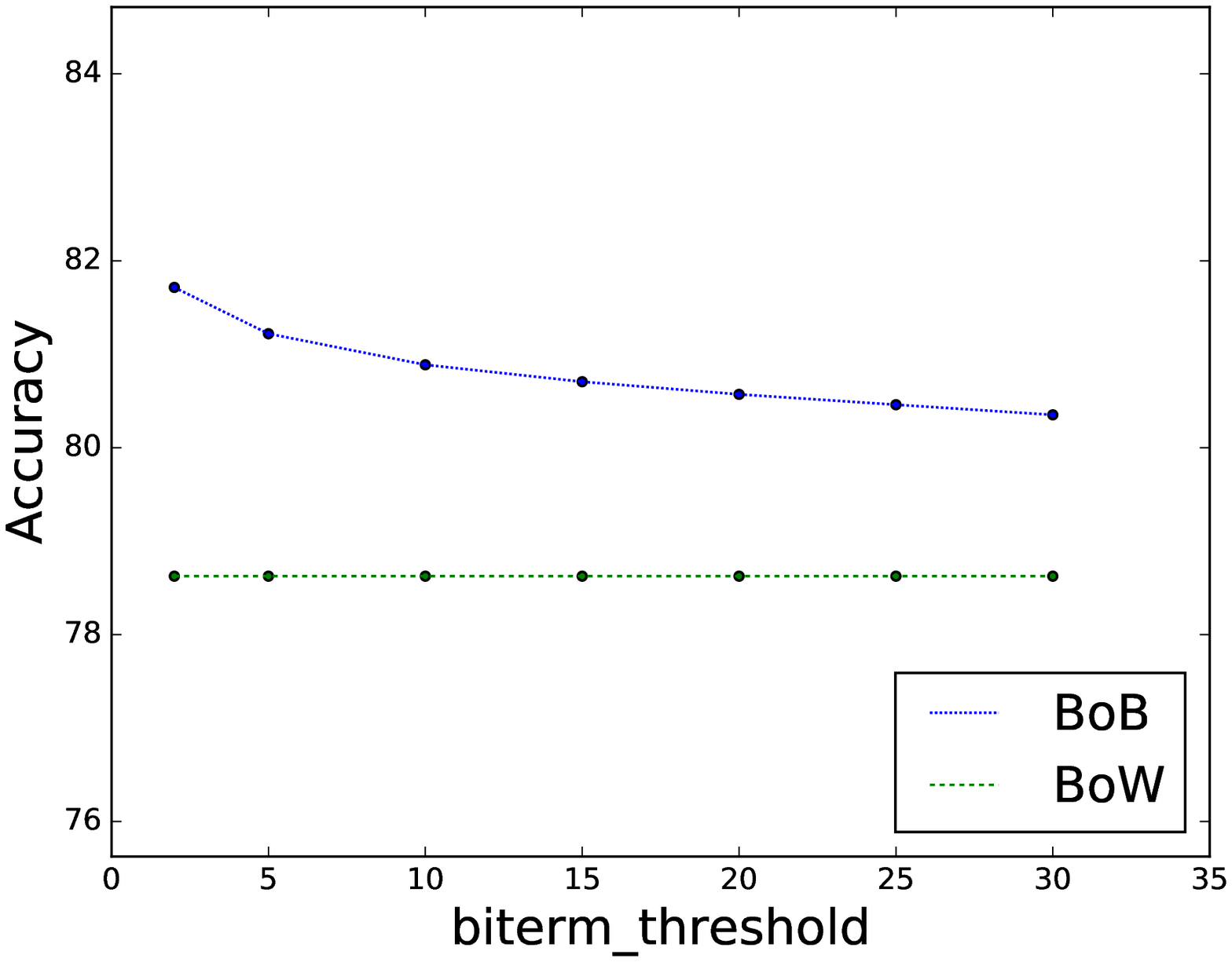}}
\end{center}
\caption{Classification performance on \textit{Yahoo Questions} and \textit{Nytimes Titles} with the weighting schema \textit{tf-idf} and the document representations BoB and BoW.}
\label{fig:classification_tfidf}
\end{figure*}
\begin{figure*}
\begin{center}
\subfloat[Ohscal]{\includegraphics[scale=0.3]{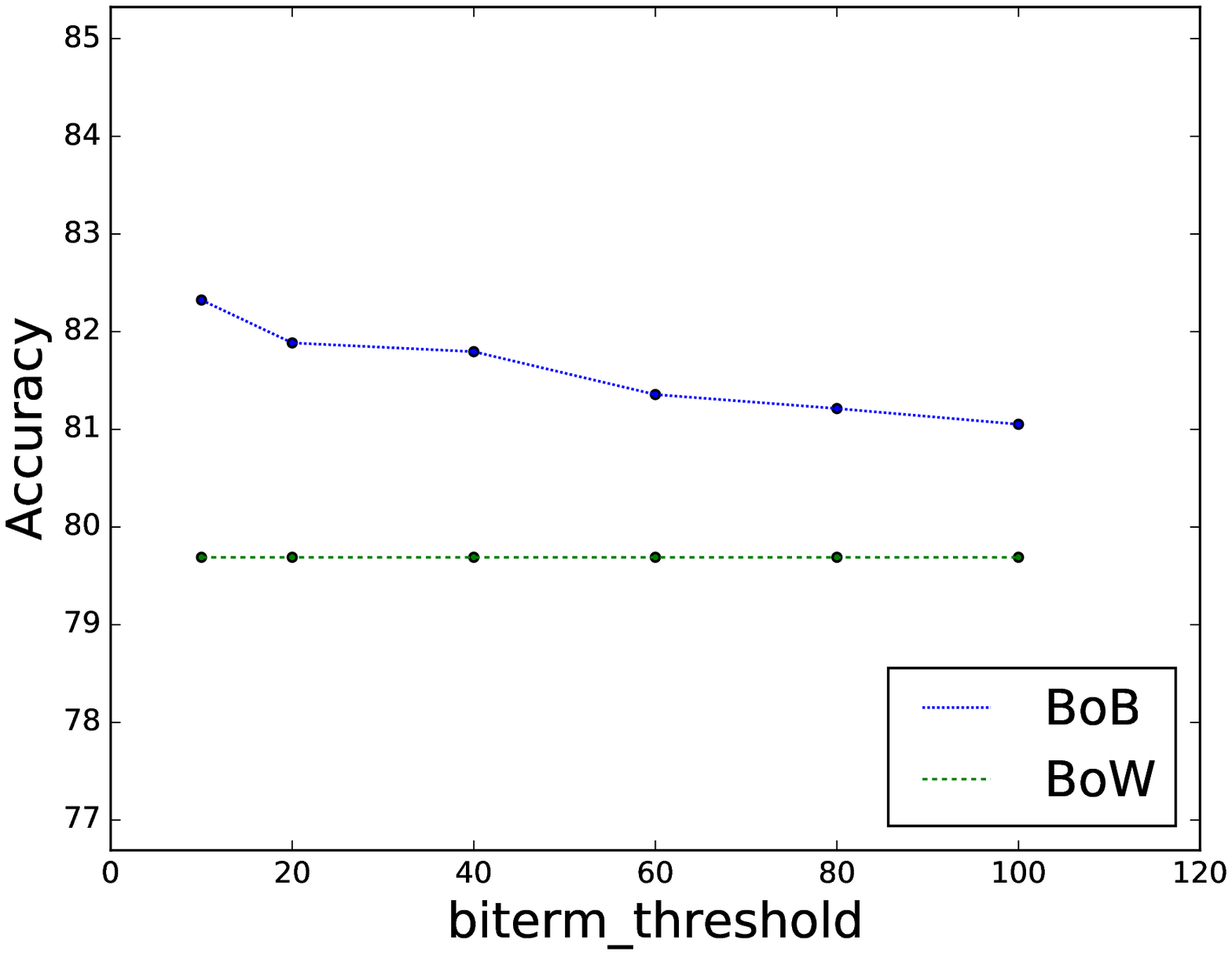}}
\subfloat[20Newsgroup]{\includegraphics[scale=0.3]{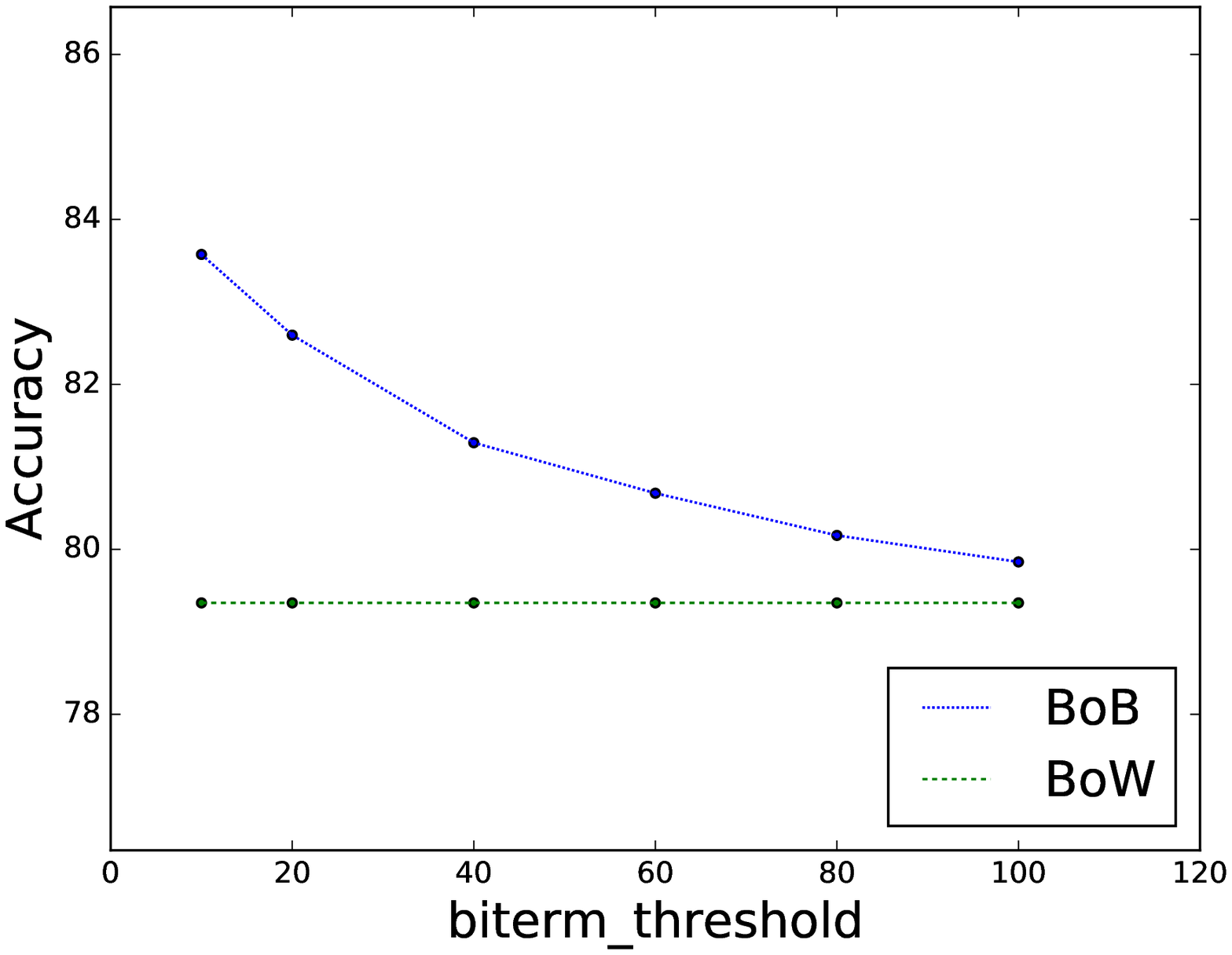}}
\end{center}
\caption{Classification performance on \textit{Ohscal} and \textit{20Newsgroup} with the weighting schema \textit{tf} and the document representations BoB and BoW.}
\label{fig:classification_long_tf}
\end{figure*}

\begin{figure*}
\begin{center}
\subfloat[Ohscal]{\includegraphics[scale=0.3]{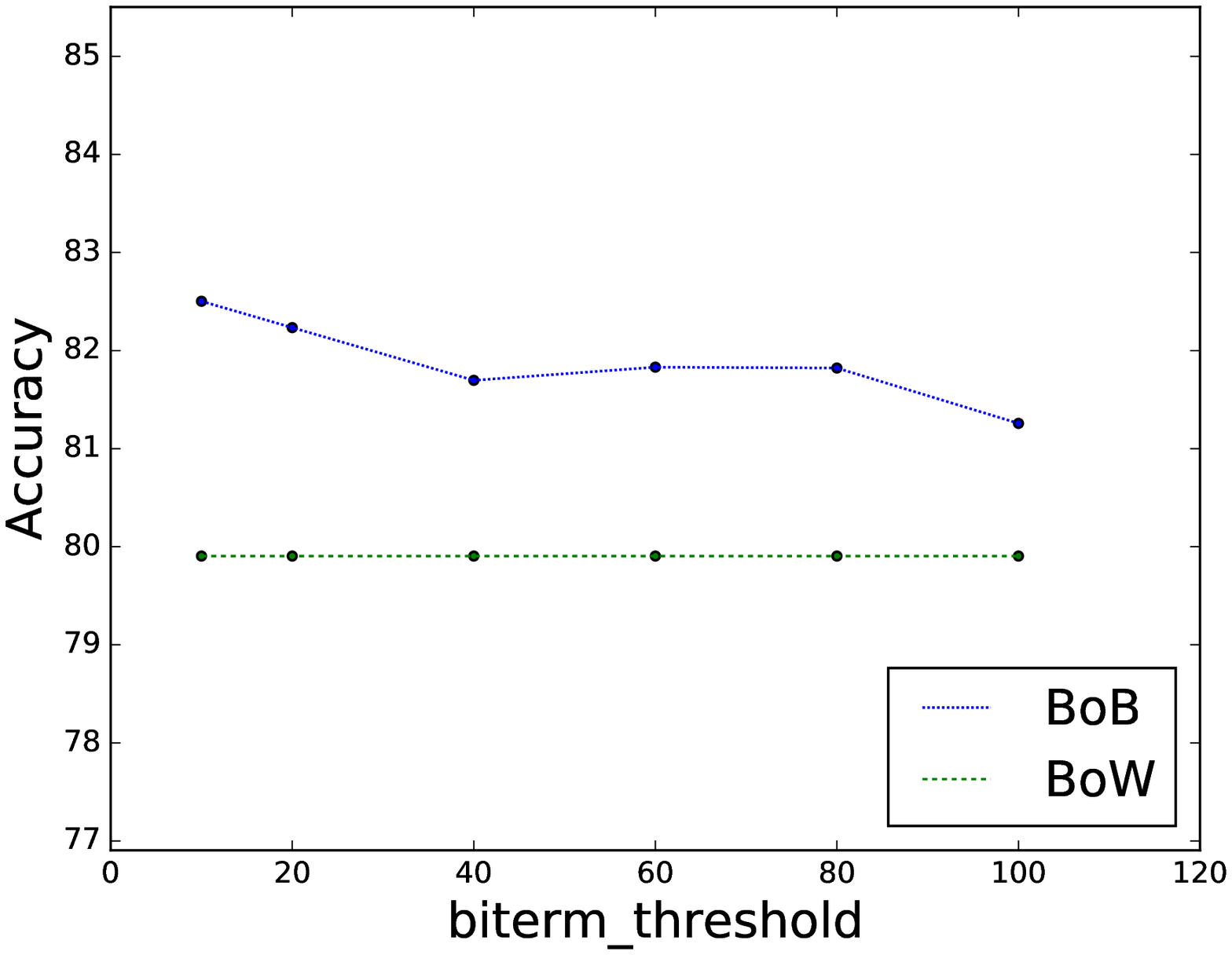}}
\subfloat[20Newsgroup]{\includegraphics[scale=0.3]{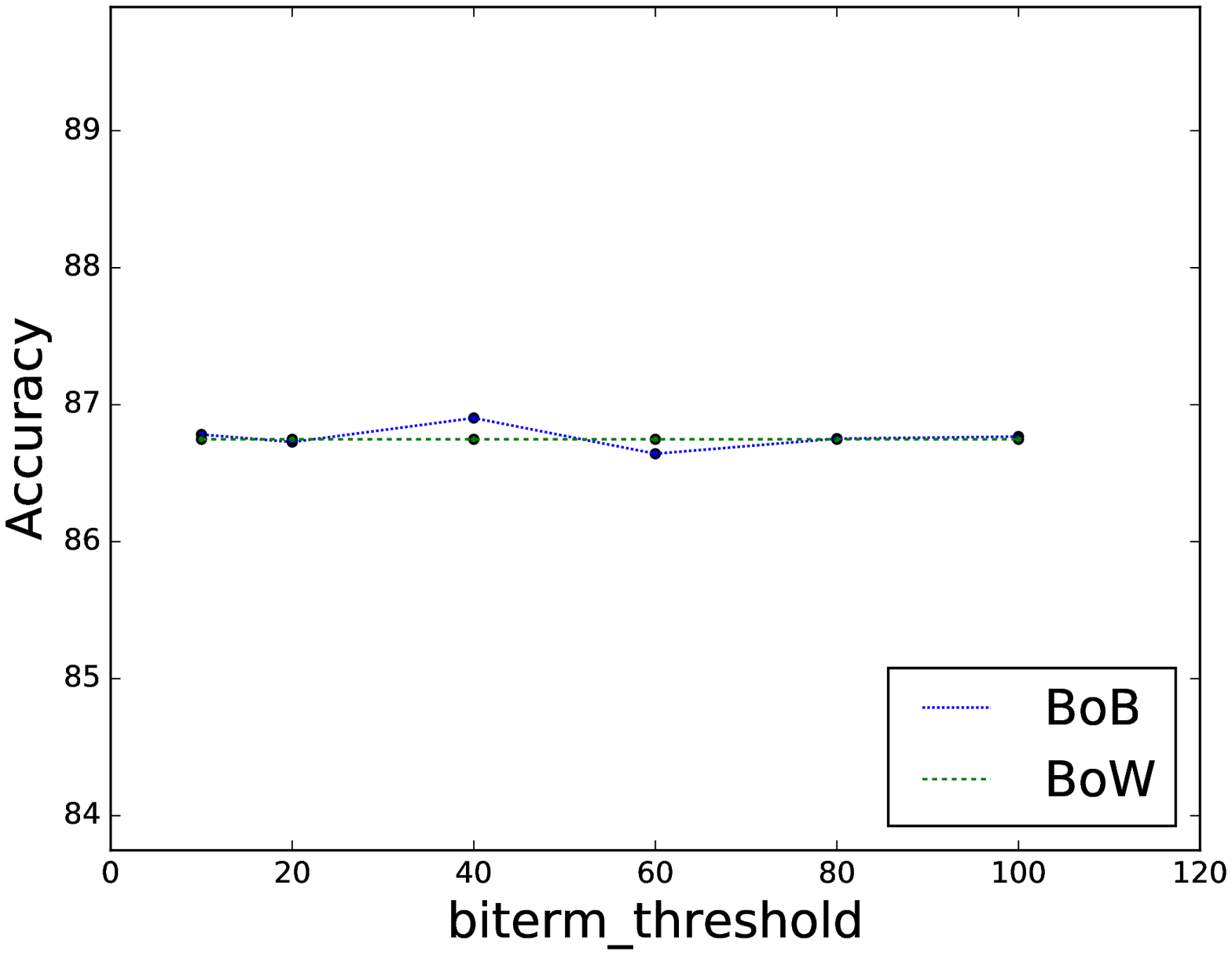}}
\end{center}
\caption{Classification performance on \textit{Ohscal} and \textit{20Newsgroup} with the weighting schema \textit{tf-idf} and the document representations BoB and BoW.}
\label{fig:classification_long_tfidf}
\end{figure*}
\par
\textbf{Settings: } For BoB, we examine different vocabulary sizes $V_b$ regulated by changing the biterm threshold. This help to see the influence of $V_b$ on the classifiers. Due to the large size of vocabulary in normal texts, we use high biterm thresholds to make the implementation of BoB more practical. The changing schedule of $V_b$ is described in Table \ref{tab:classification} (for short texts) and Table \ref{tab:long_vocabulary_biterm} (for normal texts). The statistics about the average lengths of the BoB document representation with respect to different biterm thresholds are described in Table \ref{tab:long_doc_leng_biterm}.
\begin{table}[t]
\centering
\caption{Vocabulary size ($V_b$) in BoB of Yahoo Questions and Nytimes Titles.}
\label{tab:classification}
\resizebox{\textwidth}{!}{
\begin{tabular}{|c|c|c|c|c|c|c|c|}
\hline
Biterm threshold & 2 & 5 & 10 & 15 & 20 & 25 & 30  \\ \hline
Yahoo Questions & 722,238&216,322&100,671&68,348&54,016&45,971&40,926\\ \hline
Nytimes Titles &  2,455,045 & 759,700 & 348,613 & 228,130 & 172,325 & 140,842&121,263 \\ \hline
\end{tabular}
}
\end{table}

\begin{table}
\centering
\caption{Vocabulary size ($V_b$) in BoB of 20Newsgroup and Ohscal.}
\label{tab:long_vocabulary_biterm}
\begin{tabular}{|c|c|c|c|c|c|c|}
\hline
biterm threshold &10&20&40&60&80&100 \\ \hline
$V_b$ in 20Newsgroup&2,508,921&890,744& 306,865& 172,881& 122,898&99,096 \\ \hline
$V_b$ in Ohscal&410,103&186,962& 81,751&49,978& 35,800& 28,105\\ \hline
\end{tabular}
\end{table}

\begin{table}
\centering
\caption{Average length of document representation in BoB of \textit{20Newsgroup} and \textit{Ohscal}}
\label{tab:long_doc_leng_biterm}
\begin{tabular}{|c|c|c|c|c|c|c|}
\hline
biterm threshold &10&20&40&60&80&100 \\ \hline
20Newsgroup&2838.14&1772&990&670.54&499.75&394.19 \\ \hline
 Ohscal&1144.22&877.30&622.08&485.45&398.44&337.39\\ \hline
\end{tabular}
\end{table}
\begin{table}
\centering
\caption{Execution time (in seconds) of 5-fold Linear SVM on the datasets \textit{Ohscal} and \textit{20Newsgroup} with BoW and BoB and different biterm thresholds.}
\label{tab:time_long}
\begin{tabular}{|c|c|c|c|c|c|c|c|}
\hline
dataset&bow       & 10      & 20     & 40     & 60     & 80      & 100   \\ \hline
Ohscal &2.61   & 149.09 & 84.76  & 51.22  & 36.41  & 27.79   & 8.64  \\ \hline
20Newsgroup&20.13 & 2163.35 & 720.52 & 282.65 & 153.66 & 104.04 & 31.81 \\ \hline
\end{tabular}
\end{table}
\begin{table}
\centering
\caption{Execution time (in seconds) for classification in BoW and BoB (with different biterm thresholds).}
\label{tab:time_shorttext}
\resizebox{\textwidth}{!}{
\begin{tabular}{|c|c|c|c|c|c|c|c|c|}
\hline
biterm\_threshold & bow        & 2           & 5           & 10          & 15         & 20         & 25         & 30         \\ \hline
Yahoo Questions   & 74.77  & 168.84  & 135.00  & 119.84  & 112.59 & 103.67 & 99.86  & 97.94  \\ \hline
Nytimes Titles            &705.41 & 2426.15& 1401.13 & 1171.33& 962.71 & 944.27 & 923.17 &  737.08 \\ \hline
\end{tabular}
}
\end{table}

\textbf{Experimental result:}

\par 
Figure \ref{fig:classification_tf} and Figure \ref{fig:classification_tfidf} show the results of classification on the $2$ short-text datasets \textit{Yahoo Questions} and \textit{Nytimes Titles} respectively. In general, the accuracy of BoB is always higher than the accuracy of BoW for both weighting schemas \textit{tf} and \textit{tf-idf}. This further demonstrates the effectiveness of BoB for the problem of short text classification. We also see that the weighting schema \textit{tf-idf} has better accuracy than \textit{tf} in both BoB and BoW, suggesting that the global information from corpus (i.e, the document frequency) provides useful evidences for the SVM classifiers.

\par 
Figure \ref{fig:classification_long_tf} and Figure \ref{fig:classification_long_tfidf} report the classification performance on the $2$ normal-text datasets for both BoB and BoW. The observation for the \textit{Ohscal} dataset is very similar to the short text datasets in that BoB significantly outperforms BoW with both \textit{tf} and \textit{tf-idf form} for any value of biterm thresholds. Moreover, the smaller the biterm threshold is, the more biterms are kept, and the better the classification performance of BoB is. For \textit{20Newsgroup}, BoB only outperforms BoW with the weighing schema \textit{tf} and is only comparable to BoW with \textit{tf-idf}. Note that the average length of documents in \textit{20Newsgroup} (79.97) is greater than that of \textit{Ohscal} (i.e, 60.42). These evidences suggest that BoB performs better than BoW for short and medium texts, and performs at least as well as BoW for long texts.
\par
\textbf{Execution time in BoB and BoW.}
\par 
Table \ref{tab:time_shorttext} and \ref{tab:time_long} report the training time for $4$ datasets in classification task. For \textit{Yahoo Questions} and \textit{Nytimes Titles} (short text collections), execution time might not be a critical issue when we set high biterm threshold. For example, as threshold is 30, the execution time of BoB is not much higher than BoW while the performance of BoB outperforms BoW. For \textit{Ohscal} (medium text collection) and \textit{20Newsgroup} (long text collection), there is a trade-off between accuracy and execution time when applying BoB. However, the models can be trained only once in practice and BoB is still a practical representation in the test time. 

\textbf{Comparing Biterm with Bigram} 

Bigram is a common representation method which also models two words co-occurring in a document. However, the difference between bigrams and biterms is that bigram captures the word order to capture the meaning of texts. In this part, we evaluate Bag of Biterm and Bag of Bigram when using with SVM. We conduct experiments on the two datasets: \textit{TMNTitle} (short-text), and \textit{20Newsgroup} (normal-text). 

Table \ref{tab:threshold_bigram} and Table \ref{tab:bigram_empty} show the size of bigram vocabulary and the number of empty docs respectively when we use different thresholds. Reminding that a bigram would be removed if the number of documents containing it is less than the the corresponding threshold. As we can see, there are many documents which are removed completely with the setting threshold equals one. Furthermore, Table \ref{tab:bigram_freq} shows the statistics about document frequency of bigrams, i.e., the number of bigrams appears in exactly $n$ documents ($n$ is the frequency). For \textit{TMNTitle} (short-text data), most of the bigrams just appear in one or two documents, while in \textit{20Newsgroup} (normal-text data), there are many bigrams appearing in at least 10 docs. This shows that bigramprovide very little context information for dealing with short texts.

\begin{table}[h]
\centering
\caption{Vocabulary size for Bag of Bigram in TMNTitle and 20Newsgroup.}
\label{tab:threshold_bigram}
\begin{tabular}{|c|c|c|c|c|c|c|}
\hline
Bigram Threshold & 1 & 2 & 5 & 10 & 20 & 30  \\ \hline
TMNTitle & 110,181 &	17,911 &	1,902 & 475 & 124 & 62\\ \hline
20Newsgroup &  1,121,444 & 434,160 & 98,753 & 31,472 & 10,014 & 5,012 \\ \hline
\end{tabular}

\end{table}

\begin{table}[h]
\centering
\caption{The number of empty docs when cutting bigram by threshold.}
\label{tab:bigram_empty}
\begin{tabular}{|c|c|c|c|}
\hline
Bigram Threshold & 1 & 2 & 5  \\ \hline
TMNTitle & 0 & 7465 & 12505 \\ \hline
20Newsgroup &  0 & 26 & 102 \\ \hline
\end{tabular}

\end{table}

\begin{table}[h]
\centering
\caption{The document frequency of bigrams in \textit{TMNTitle} and \textit{20Newsgroup}.}
\label{tab:bigram_freq}
\begin{tabular}{|c|c|c|c|c|c|c|c|c|c|c|}
\hline
Doc frequency & 1 & 2 & 3 &  4 & 5 & 6 & 7 & 8 & 9 &  10 \\ \hline
TMNTitle & 92270 & 11912 & 2957 & 1140 & 616 & 326 & 217 & 159 & 109 & 78 \\ \hline
20Newsgroup & 687284 & 207584 & 83713  & 44110 & 24622 & 16831 & 11341 & 8129 & 6358 & 4667 \\ \hline
\end{tabular}

\end{table}

\begin{figure*}[t]
\begin{center}
\includegraphics[scale=0.3]{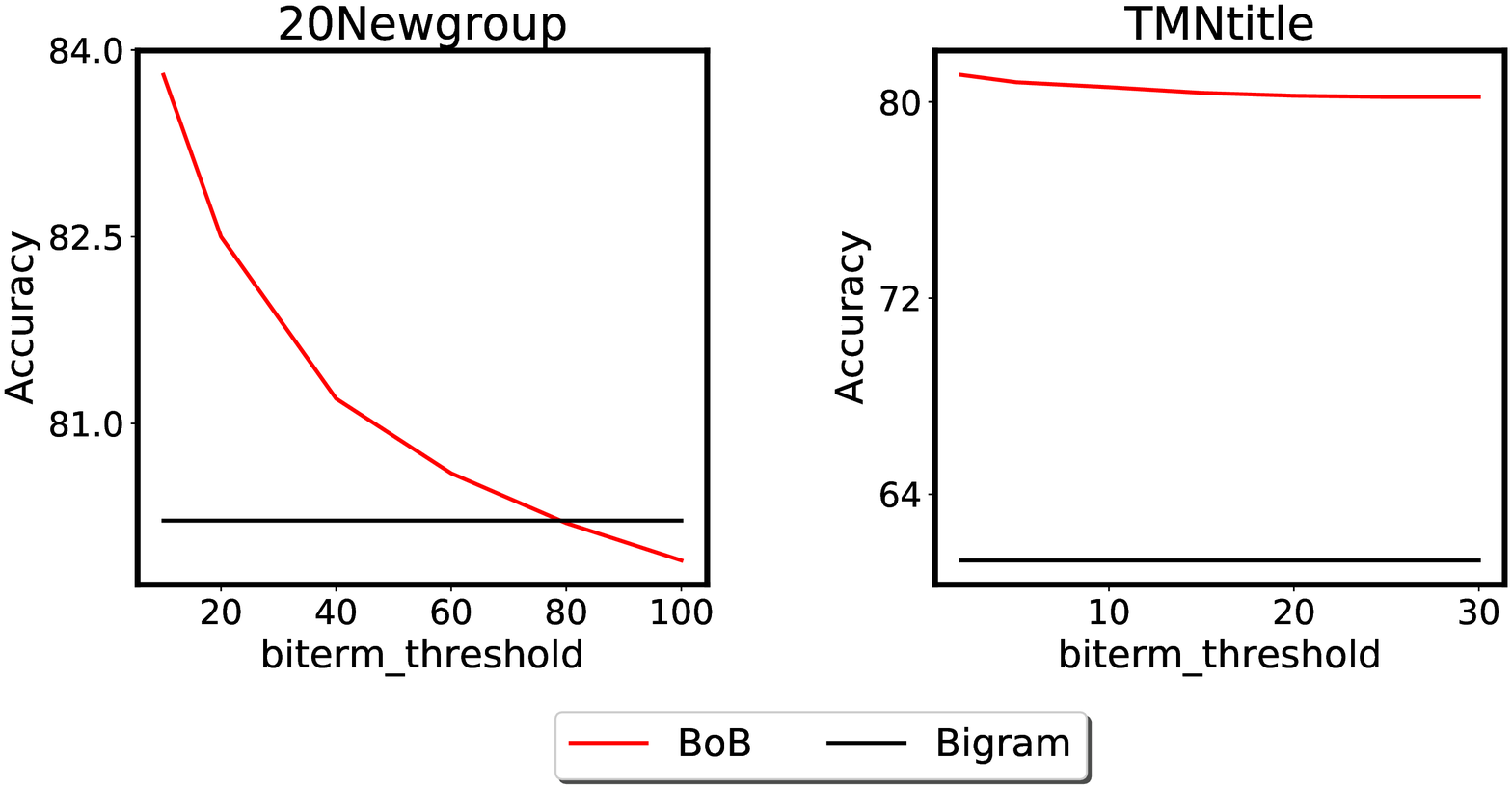}
\end{center}
\caption{Classification performance on \textit{20Newsgroup} and \textit{TMNTitle} with the weighting schema \textit{tf} and the document representations BoB and BoW.}
\label{fig:bigram_tf}
\end{figure*}

\begin{figure*}[t]
\begin{center}
\includegraphics[scale=0.3]{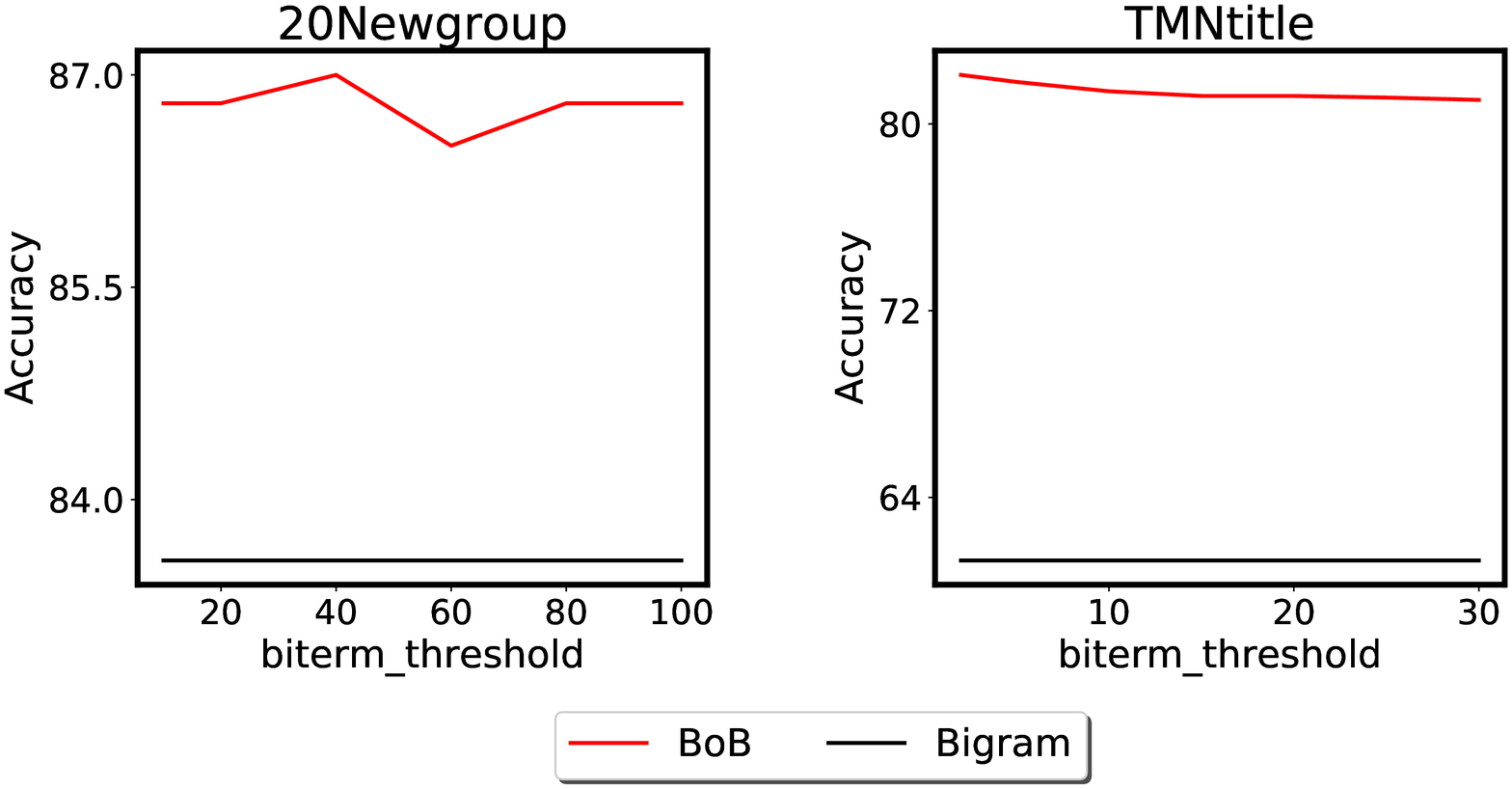}
\end{center}
\caption{Classification performance on \textit{20Newsgroup} and \textit{TMNTitle} with the weighting schema \textit{tf-idf} and the document representations BoB and BoW.}
\label{fig:bigram_tfidf}
\end{figure*}

Figure \ref{fig:bigram_tf} and \ref{fig:bigram_tfidf} show the classification performance with the \textit{tf} and \textit{tf-idf} settings. We choose the bigram threshold setting which gains the best performance for the bigram models, i.e., threshold equals two. The performance of Bigram is worse than those of BoB in both datasets. The reason might be that bigram does not contain much co-occurrence information in short texts which is very important in topic models.

\subsection{Evaluation of BBM}
In this section, we evaluate the performance of BBM on two tasks: (1) we compare HDP-B with the base HDP model with both the BoB and BoW representations, and (2) we investigate the effectiveness of LDA-B in the two contexts of online and streaming data.

\subsubsection{Evaluation of online HDP-B}
In order to evaluate the effectiveness of BBM, this section compares (online) HDP-B with the primitive HDP when either BoB or BoW is used as the representation.

\par 
\textbf{Baseline methods:} We use the online HDP as the implementation of the base topic model HDP in this section. Online HDP is a standard method in the online environment that has been improved significantly with the BoB representation (i.e, Section \ref{sub:bob-topicModel}).

\par
\textbf{Settings:} The parameters are chosen according to those in \cite{mai2016enabling}. In particular, the learning rate parameters $\tau$ and $\kappa$ form a grid: $\tau \in \{1, 20, 40, 60, 80, 100\}$, $\kappa \in \{0.6, 0.7, 0.8, 0.9\}$. The result is averaged over the values of the 24 LPPs associated with the 24 settings. 

\par 
\textbf{Experimental Result:}

\begin{figure*}[t]
\begin{center}
\includegraphics[scale=0.35]{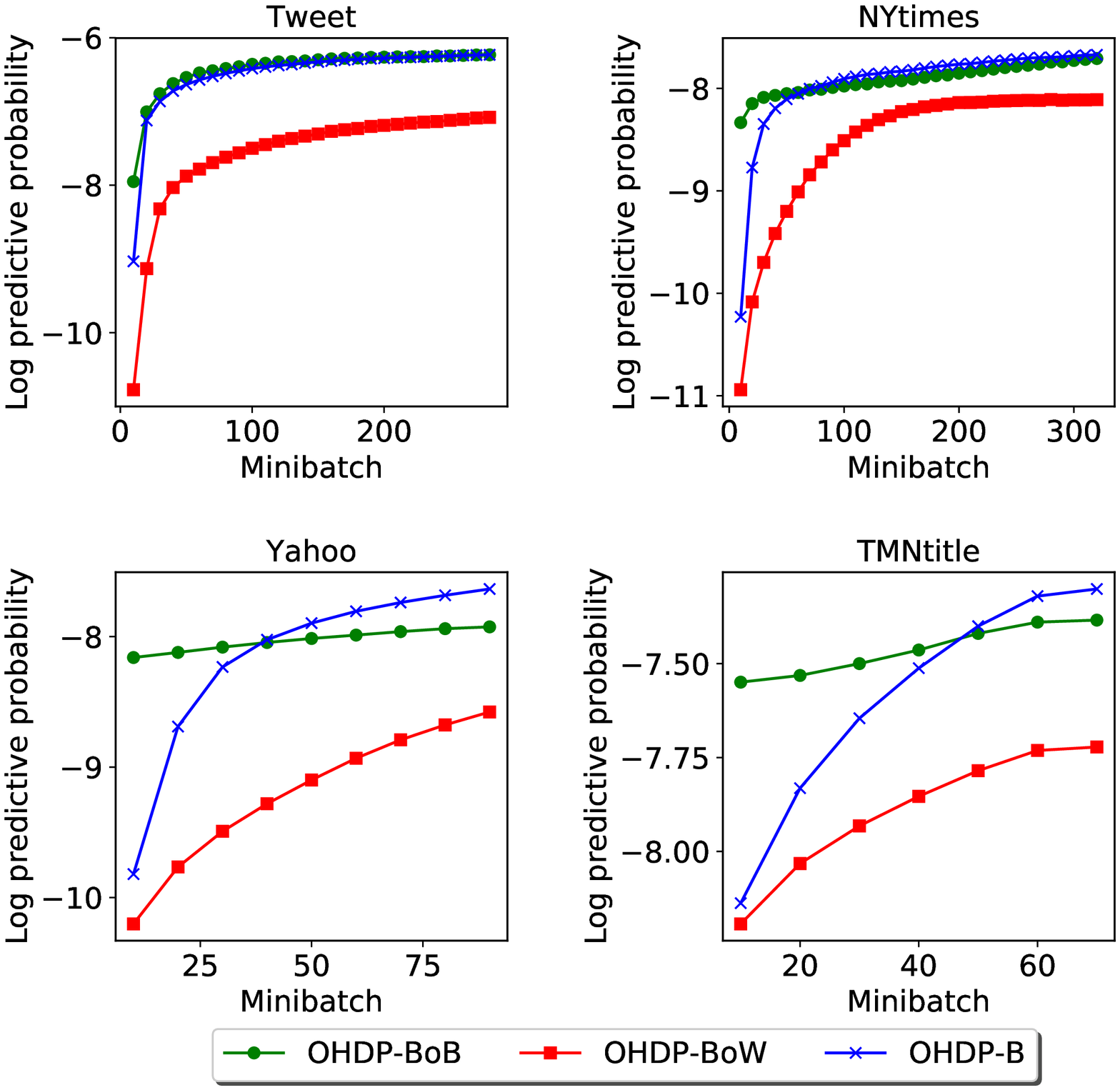}
\end{center}
\caption{Performance of online HDP-B and online HDP (with BoW and BoB). The higher is the better}
\label{fig:BBM_hdp}
\end{figure*}

The performance of the models is shown in Figure \ref{fig:BBM_hdp}. From this figure, we can see that the online HDP-B and online HDP with the BoB representations always performs substantially better than the online HDP that utilizes BoW as the representation. Comparing online HDP-B and online HDP with the BoB representation, we see that online HDP-B is comparable to online HDP over the \textit{Tweet} and \textit{NYtimes} datasets, and perform much better than online HDP over the \textit{Yahoo} and \textit{TMNtitle} datasets when the model receives more data (i.e, the batch size increases). Note that BBM is also an efficient method for short texts due to its mechanism to reduce the size of the vocabulary for BoB, as discussed in Section \ref{sec:III}. Such efficiency and effectiveness make BBM a practical method for short texts.

\begin{figure*}
\begin{center}
\includegraphics[scale=0.35]{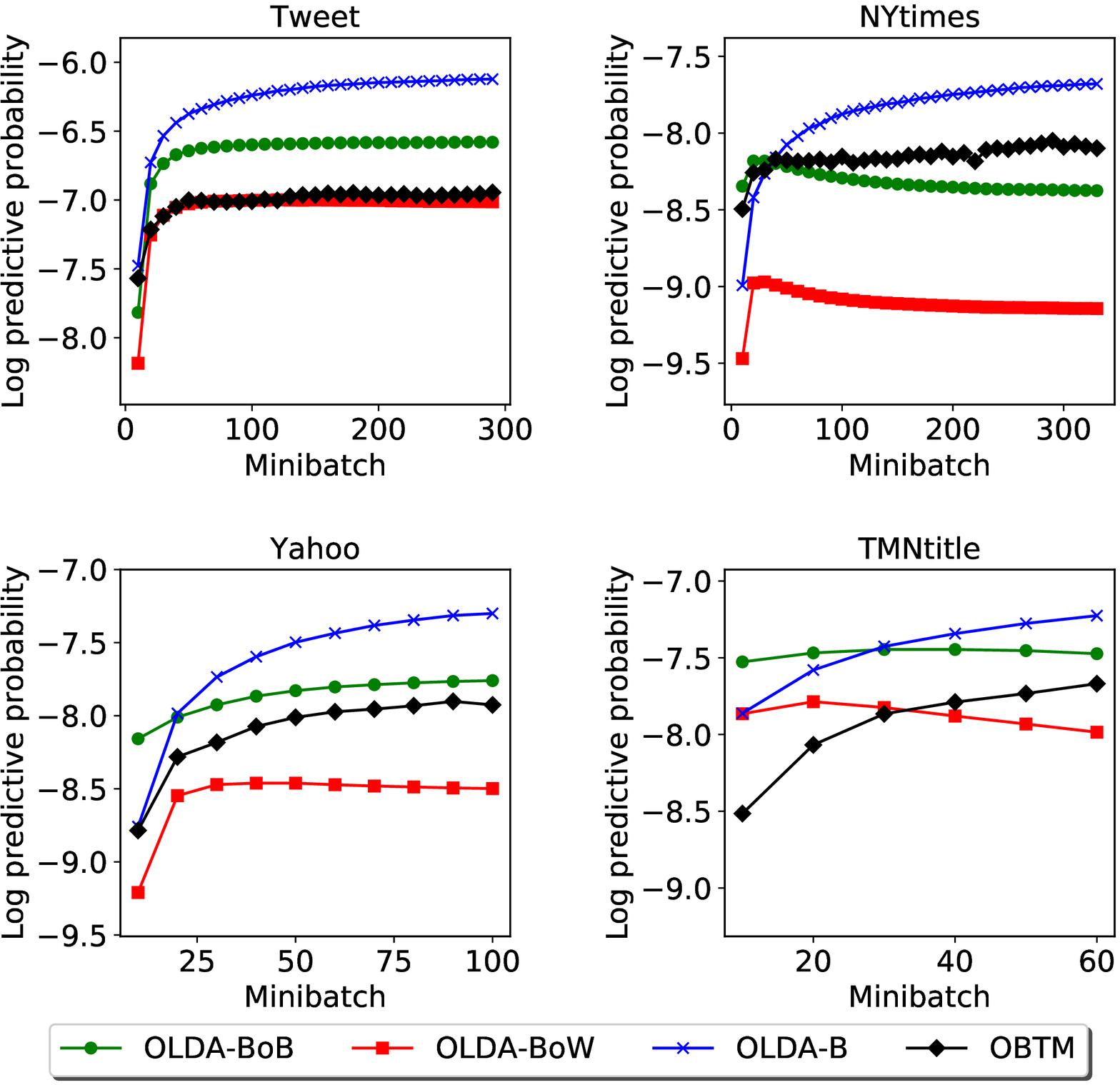}
\end{center}
\caption{Performance of online LDA-B, Online BTM, and Online LDA (with BoB and BoW). The higher is the better.}
\label{fig:lda_b}
\end{figure*}
                                                                                                                                                                                                                                                                 
\subsubsection{Evaluation of online LDA-B}
In what follows, we demonstrate the advantages of LDA-B over Biterm Topic Model (BTM) \cite{yan2014btm}, a state of the art model for short texts in online settings. Furthermore, we also compare the online LDA-B (OLDA-B) with the primitive which uses BoW and BoB as text representation methods. The learning algorithm for OLDA-B is described in Algorithm \ref{algo:SVI_LDA}.
\par 
\textbf{Baseline methods}:
\begin{itemize}
\item OBTM (Online BTM): the state-of-the-art framework for short texts that models biterms directly in the whole corpus. 
\item OLDA (Online LDA): the standard online method that is derived by applying Stochastic Variational Inference (SVI) to LDA model.
\end{itemize}
\par
\textbf{Settings}: We set the number of topics K equal to 100 for \textit{Tweet}, \textit{NYtimes} and \textit{Yahoo}, while $K=50$ for \textit{TMNtitle}. The learning rate parameters $\tau$ and $\kappa$ form a grid: $\tau \in \{1, 20, 40, 60, 80, 100\}$, $\kappa \in \{0.6, 0.7, 0.8, 0.9\}$. The result is averaged over the values of the 24 LPPs associated with the 24 settings. The parameters for Online BTM are the best values in \cite{yan2014btm}
\par 
\textbf{Experimental Result}
\par 
The results are shown in Figure \ref{fig:lda_b}. We find that OLDA-B consistently higher than the all the baseline methods. When the batch size increases (more data is provided), the performance of OLDA-B is improved dramatically while OBTM and OLDA with BoB only have insignificant increases or remains stable. An explanation for this phenomenon is that when much more data arrives (in OBTM), the assumption of aggregating all documents into a single one makes the corpus ambiguous, and possibly affects the performance. In contrast, BBM infers and extracts latent topics at the document level, thus improving the performance when more data is received. 
\subsubsection{Evaluation of streaming LDA-B}
In this part, we demonstrate the advantages of LDA-B in the context of streaming data. 
\par 
\textbf{Baseline methods}:
\begin{itemize} 
\item SVB (Streaming Variational Bayes)\cite{broderick2013streaming} a framework that enables LDA  to work in the streaming environment.
\item KPS (Keeping priors in Streaming Learning)\cite{duc2017keeping} a streaming framework that incorporates the prior knowledge induced from word embedding.
\end{itemize}
These methods will function as the baselines for the proposed BBM method in this section.
\par
\textbf{BBM models for the streaming data}: We consider the following two applications of BBM to the domain of streaming data, i.e, SVB-B and KPS-B. SVB-B (as in Algorithm 3) is a streaming method that is derived by applying the SVB framework \cite{broderick2013streaming} on the LDA-B model. KPS-B (as in Algorithm 4) is derived in a similar way except that the SVB framework is replaced by the KPS framework \cite{duc2017keeping} in the process. We will compare these methods with the baselines to evaluate the effecitiveness of BBM.
\par 
\textbf{Prior in use:} For KPS and KPS-B, we use word embeddings as a prior knowledge. In particular, we employ the word embeddings provided by Pennington et al. (2014)\footnote{\url{https://nlp.stanford.edu/projects/glove/}} that was trained on 6 billion tokens of the whole Wikipedia 2014 and Gigaword 5 corpus . In KPS and KPS-B, the dimensionality of word embeddings is equal to the number of topics K, with $K=100$ for \textit{Tweet}, \textit{NYtimes} and \textit{Yahoo} datasets, and $K=50$ for \textit{TMNtitle}. We also normalize the word embeddings by the softmax function to guarantee all the dimensions are non-negative.

\par
\textbf{Settings:} In these experiments, the number of topics $K$ is set to 100 for \textit{Tweet}, \textit{NYtimes} and \textit{Yahoo} datasets, whereas $K=50$ for \textit{TMNtitle}. For SVB and SVB-B, the parameters are inherited from the best ones for SVB in \cite{broderick2013streaming}. In particular, both $\alpha$ and $\eta$ are set to 0.01. For KPS and KPS-B, $\alpha$ is also set to 0.01, and $\eta$ is set by the word embeddings.

\par 
\textbf{Experimental Result}

\begin{figure*}[t]
\begin{center}
\includegraphics[scale=0.24]{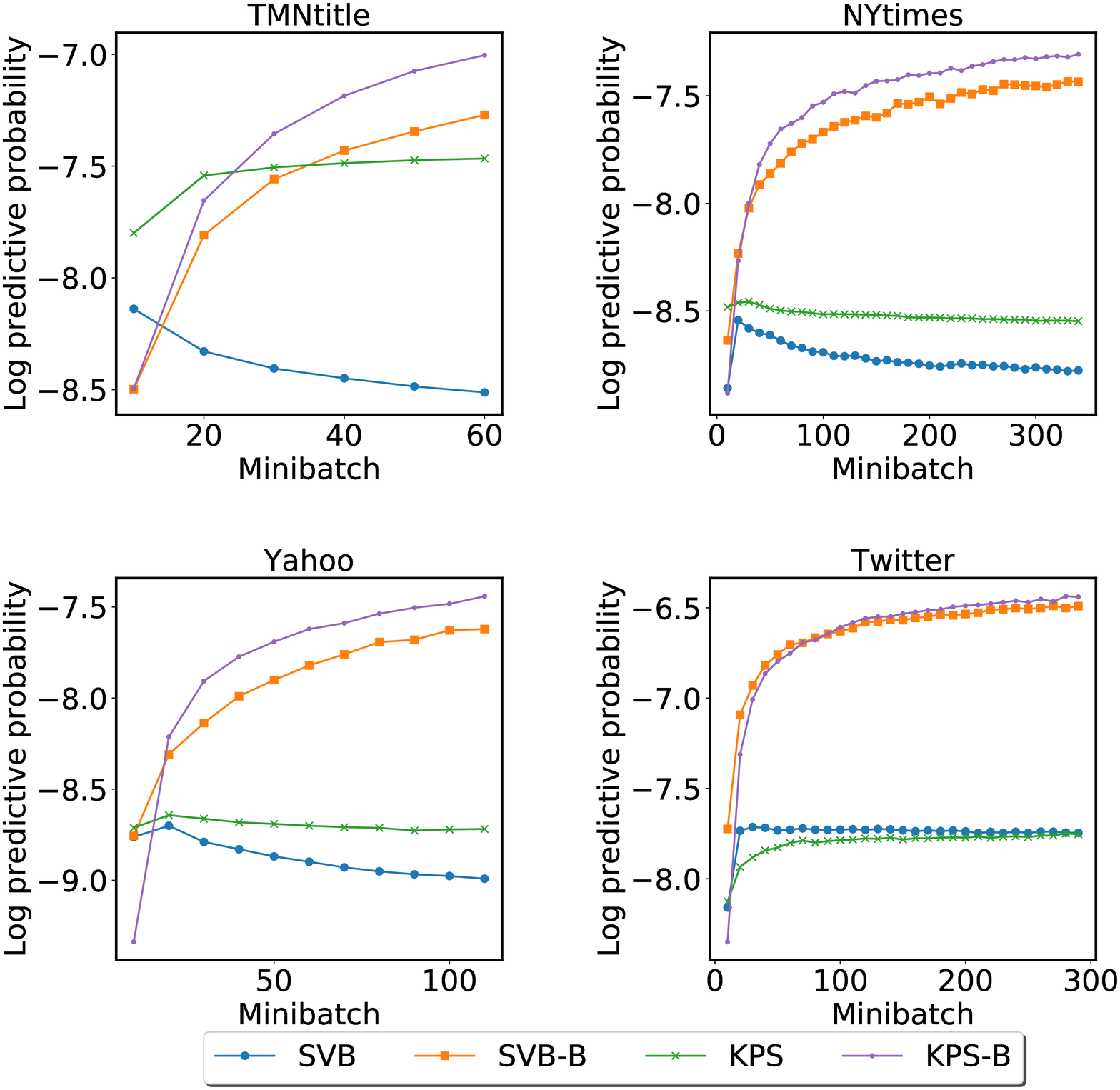}
\end{center}
\caption{Performance of $4$ streaming methods: SVB-B, KPS-B, SVB and KPS. The higher is the better.}
\label{fig:bbm_streaming}
\end{figure*}

Figure \ref{fig:bbm_streaming} shows the performance of the models for the streaming data on the four short text datasets. It is clear from the figure that that both SVB-B and KPS-B performs significantly better than all the baseline models. Furthermore, we see that when more data is provided (i.e, the batch size increases), the performance of SVB-B and KPS-B is improved dramatically while KPS only has an insignificant improvement at the beginning and remains stable afterward. It shows that BBM can improve the primitive significantly in the context of streaming. In addition, comparing SVB-B and KPS-B, we see that KPS-B almost always performs better than SVB-B, except for \textit{Twitter}. One reason is that the prior might not be relevant to the corpus of social networks like \textit{Twitter}. Nevertheless, it indicates that exploiting human knowledge for BBM can help to improve the performance of the models. 
\begin{figure*}
\begin{center}
\includegraphics[scale=0.24]{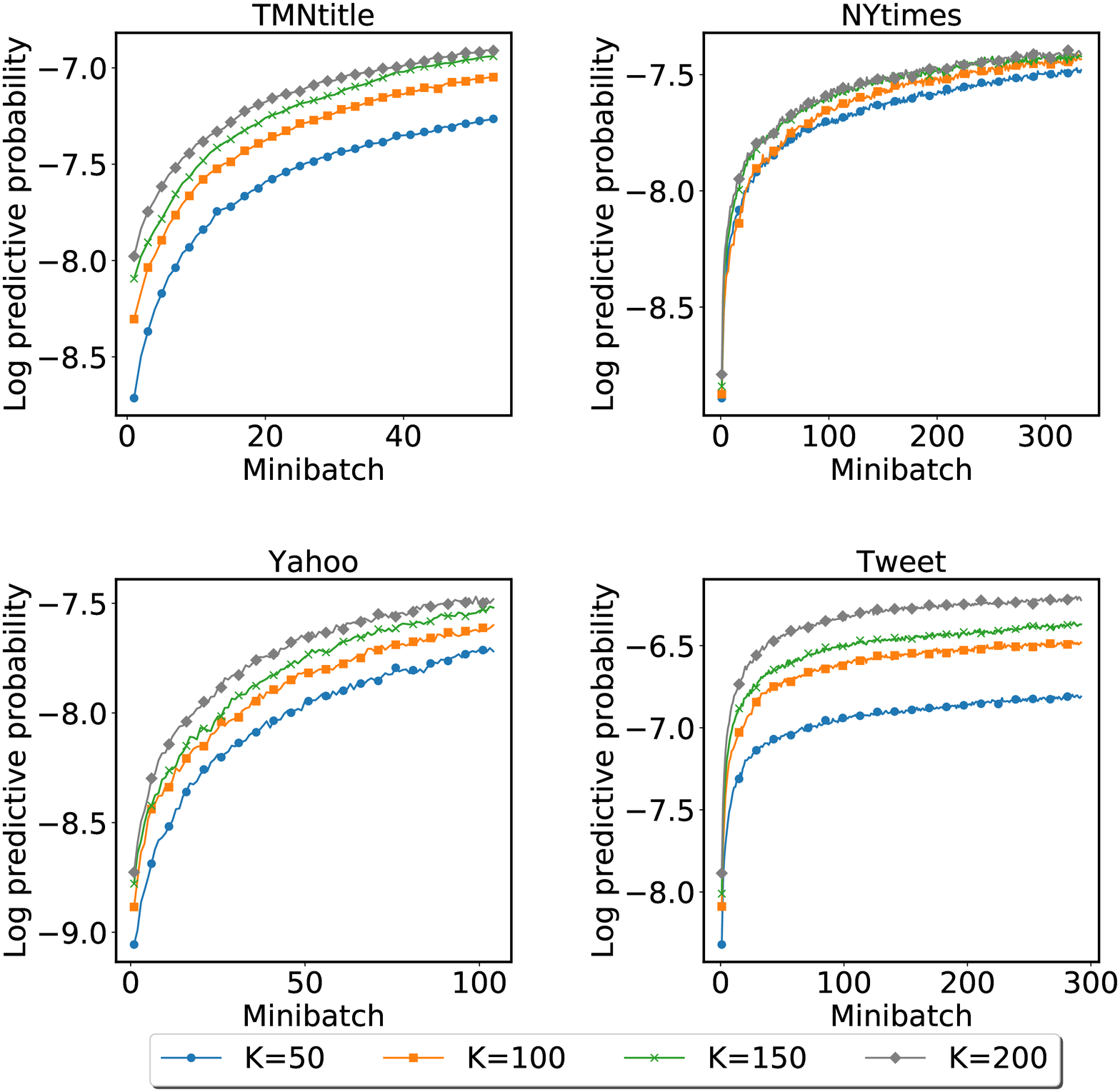}
\end{center}
\caption{The sensitivity of SVB-B with different number of topics}
\label{fig:sensitive_topics}
\end{figure*}
\begin{figure*}
\begin{center}
\includegraphics[scale=0.24]{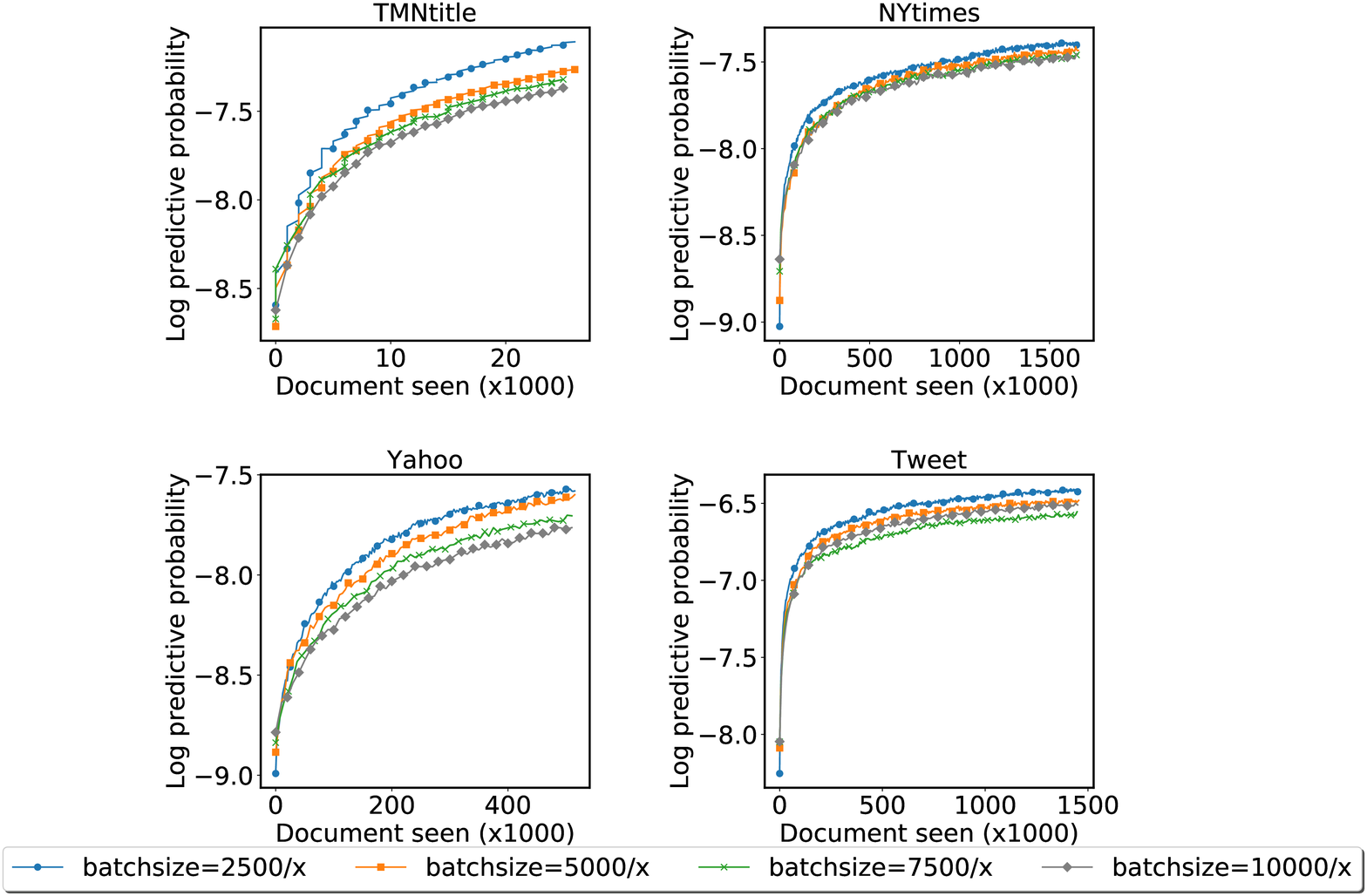}
\end{center}
\caption{The sensitivity of SVB-B with different batchsize ($x=10$ for $TMNtitle$ and $x=1$ for the other datasets)}
\label{fig:sensitive_batchsize}
\end{figure*}

\par 
\textbf{Parameter sensitivity}
\par
Finally, we investigate the effects two parameters when their values are varied, i.e, the number of topics $K$ and the batch size. Figure \ref{fig:sensitive_topics} and \ref{fig:sensitive_batchsize} show the predictive accuracy of SVB-B when regulating these two parameters. Note that in Figure \ref{fig:sensitive_batchsize}, the batch size is in the range of $\{2500, 5000, 7500, 10000\}$ for three datasets: \textit{NYtimes}, \textit{Yahoo} and \textit{Tweet}, and $\{250, 500, 750, 1000\}$ for TMNtitle due to the minor size of the dataset. As we can see from Figure \ref{fig:sensitive_batchsize}, in general, the higher values of $K$ yield better performance for the models than the lower values. The performance gaps between different values of $K$ are large except for the case of the \textit{NYtimes} dataset whose accuracy seems stable for different numbers of topics. One possible reason is that words used in \textit{NYtimes} are often formal and restricted in a limited range of topics. In contrast, the other datasets come from social media that often involve informal texts, leading to the extended numbers of topics $K$.  Regarding the batch size, Figure \ref{fig:sensitive_batchsize} shows that the accuracy over the four datasets is relatively stable when the batch size changes.

\section{Conclusion \& Future works}
\label{sec:VI}
In this paper, we present a new representation for short text documents, called bag of biterms (BoB) and a framework for Modeling Bag of Biterms (BBM). The extensive experiments demonstrate many advantages of BoB over the traditional bag of words (BoW) representation, making BoB more suitable for short texts. In particular, BoB helps to reduce the negative effects of shortness and reinforce the context of the documents. These properties enable BoB to work well with unsupervised topic models such as LDA, HDP and supervised classification methods such as SVM for text classification. However, BoB has limitations on time and memory due to the size of the biterm vocabulary. The proposed BBM technique help to overcome these limitations by eliminating the need for the biterm vocabulary and modeling the two words of a biterm separately in the model. We introduce two implementation of BBM based on the primitive topic models LDA and HDP. The experimental results show that these  models perform substantially better than the base topic models over several short text datasets. Besides, we conduct experiments on the streaming environment that further demonstrate the benefits of BBM for the streaming data. In particular, the performance of the current streaming frameworks for topic models is improved when BBM is employed as the topic model in these frameworks. In the future, we plan to seek the applications of BBM in more domains and to investigate techniques to combine BBM with other text analysis methods beyond topic models.
\begin{acknowledgements}
This research is supported by Vingroup Innovation Foundation (VINIF) in project code VINIF.2019.DA18, and by the Office of Naval Research Global (ONRG) under Award Number N62909-18-1-2072, and  Air Force Office of Scientific Research (AFOSR), Asian Office of Aerospace Research \& Development (AOARD) under Award Number 17IOA031.
\end{acknowledgements}

\newpage
\bibliography{main}
\bibliographystyle{ieeetr}

\newpage
\begin{appendices}
\section{Supplementary experimental result}
\begin{figure*}[h]
\begin{center}
\includegraphics[width=0.7\textwidth]{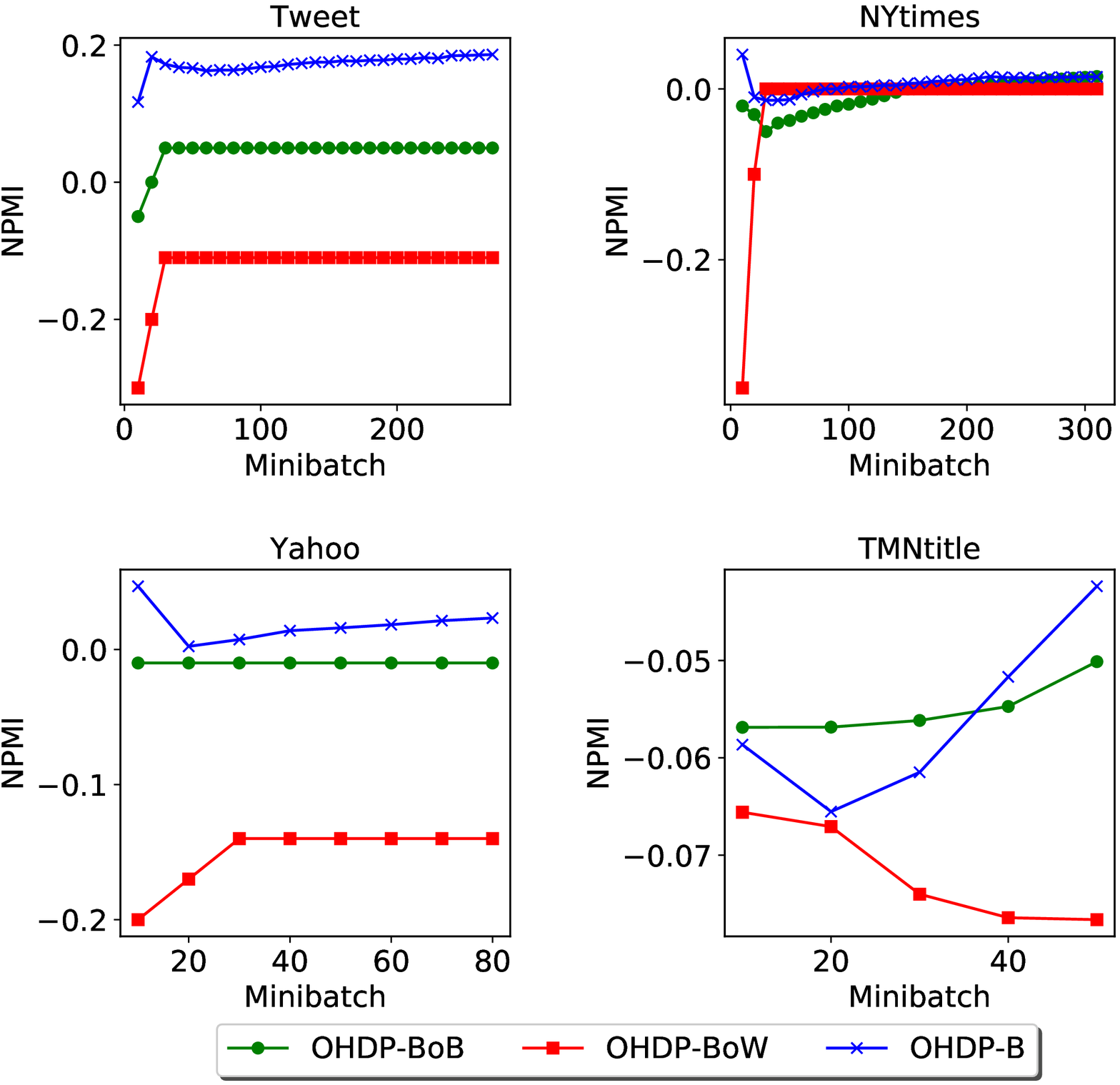}
\end{center}
\caption{The NPMI of online HDP-B and online HDP (with BoW and BoB)}
\label{fig:npmi_hdp}
\end{figure*}

\begin{figure*}[h]
\begin{center}
\includegraphics[width=0.7\textwidth]{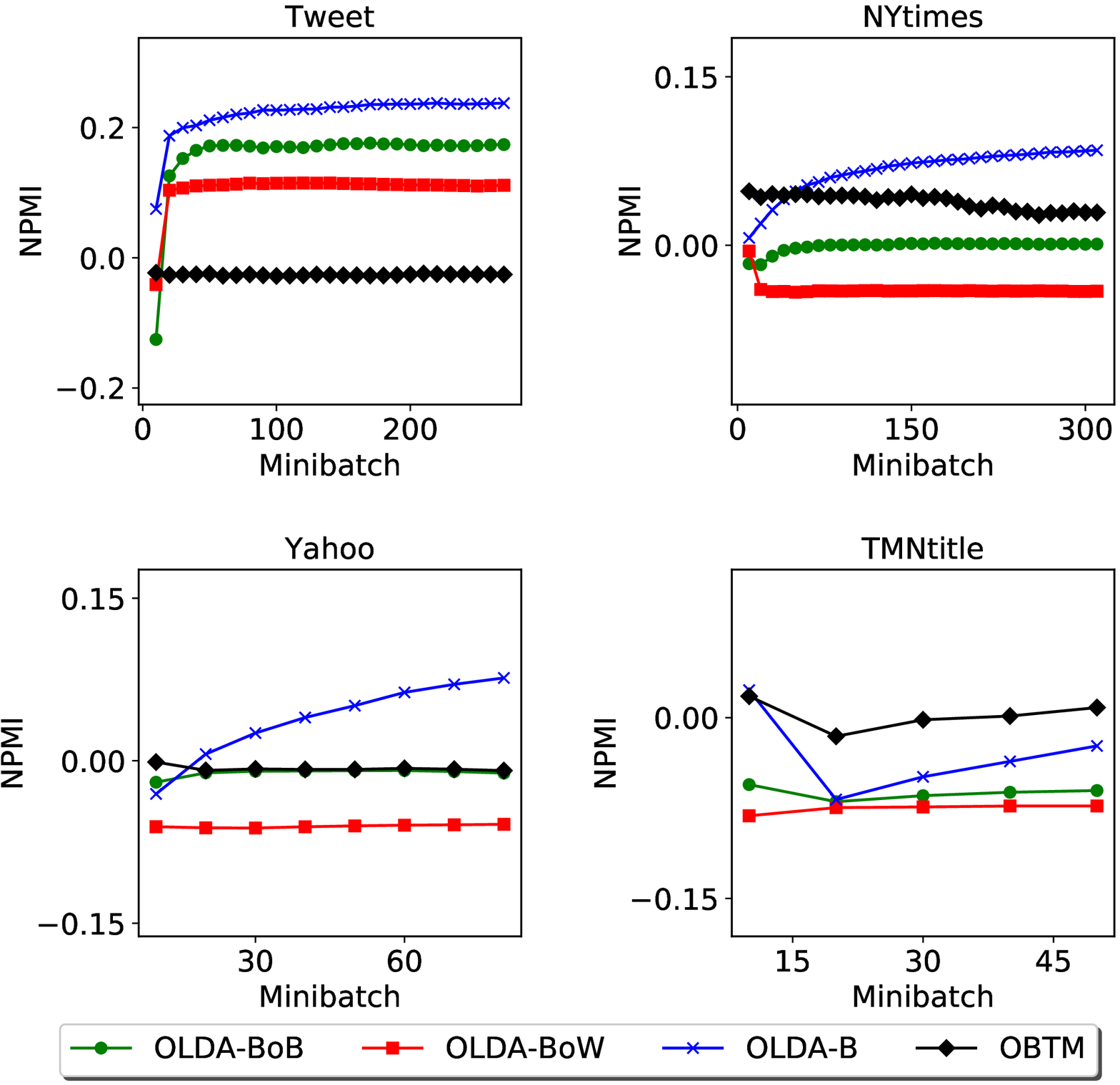}
\end{center}
\caption{The NPMI of online LDA-B, Online BTM, and Online LDA (with BoB and BoW)}
\label{fig:npmi_lda}
\end{figure*}

To strengthen the experimental result in Section \ref{sec:V}, we conduct some experiments with another evaluation metrics besides Log Predictive Probability (LPP), i.e., Normalized Pointwise Mutual Information (NPMI) \cite{bouma2009normalized}. We evaluate the NMI of the two models, Online HDP-B and Online LDA-B, compared with their base models using BoW and BoB. The settings and model in use are exactly the same as those in Section \ref{sub:bob-topicModel}. We adopt the four datasets: \textit{Tweet}, \textit{NYtimes}, \textit{Yahoo}, and \textit{TMNTitle}.

\textbf{Normalized Pointwise Mutual Information (NPMI)}: a standard metric to measure the association between a pair of discrete outcomes x and y, defined as:

\begin{equation}
\text{NPMI}(w_i) = \sum_{j}^{N-1} \frac{\log \frac{P(w_i, w_j)}{P(w_i) P(w_j)}}{-\log P(w_i, w_j)}
\end{equation}


Figure \ref{fig:npmi_hdp} and Figure \ref{fig:npmi_lda} show the evaluation of these models by NPMI. Figure \ref{fig:npmi_hdp} confirms that OHDP-B performs better than HDP in almost all of the cases. Figure \ref{fig:npmi_hdp} shows that OLDA-B achieves better result than the other three models, while OBTM ranks behind only OLDA-B in \textit{NYtimes}, \textit{Yahoo}. Also, in most of the cases in both the two figures, the results of using BoW show that this representation is not suitable with short text datasets.  
\section{Conversion of topic-over-biterms (distribution over biterms) to topic-over-words (distribution over words)}\label{appenA}
In BoB, after we finish training the model, we obtain topics that are multinomial distributions over biterms. We would like to convert these topic-over-biterms to the topics-over-words (i.e, distribution over words). Assume that $\boldsymbol\phi_k$ is a distribution over biterms of topic $k$.
The procedure to perform this conversion is as follows: \\
$
p(w_i \mid z=k)  = \sum_{j=1}^{V}p(w_i,w_j \mid z=k) 
 = \sum_{j=1}^{V}p(b_{ij} \mid z=k)
= \sum_{j=1}^{V}\phi_{kb_{ij}}
$,\\
where $V$ is the vocabulary size in BoW and $b_{ij}$ is the biterm created from the pair ($w_i,w_j$).
\par 
 As discussed in Section \ref{subsec:definition-biterm}, in the implementation of BoB, we can merge $b_{ij}$ and $b_{ji}$ into $b_{ij}$ with $i<j$. Because of the identical occurrence in every document, after finishing the training process, the value of $p(b_{ij}\mid z=k)$ will be expectedly the same as $p(b_{ji}\mid z=k)$. Therefore, in grouping these biterms into one, the conversion version of this implementation is:
$
p(w_i\mid z=k)  =  \sum_{j=1}^{V}p(b_{ij} \mid z=k) = \phi_{kb_{ii}} + \frac{1}{2}\sum_{\text{b: biterms contain } w_i}\phi_{kb}
$

\section{Parameters inference for LDA-B}\label{appenB}
\subsection{Lower bound function}
The log likelihood is bounded by the lower bound induced from the Jensen inequality:
\begin{align*}
L = \log P (\beta, \theta, z, \tilde{z} \mid \lambda, \gamma, \phi)  \geq E_{q} [ \log P(\beta, \theta, z, \tilde{z}, w \mid \eta, \alpha] - E_{q} [\log q (\beta, \theta, z, \tilde{z} \mid \gamma, \lambda, \phi) ]
\end{align*}

This lower bound can be written as follow:
\begin{align*}
L &= \sum_{k=1}^{K} E_{q} \left[ \log P (\beta_k \mid \eta_k) \right] + \sum_{d=1}^{D} E_{q} \left[ \log P (\theta_d \mid \alpha) \right] +  \sum_{d=1}^{D} \sum_{n=1}^{N_d} E_{q} \left[ \log P (z_{dn} \mid \theta_d) \right] \\ &+ \sum_{d=1}^{D} \sum_{m=1}^{M_d} E_{q} \left[ \log P (\tilde{z}_{dm} \mid \theta_d) \right] + \sum_{d=1}^{D} \sum_{n=1}^{N_d} E_{q} \left[ \log P (w_{dn} \mid z_{dn}) \right] \\ &+  \sum_{d=1}^{D} \sum_{m=1}^{M_d} E_{q} \left[ \log P (\tilde{w}^{(1)}_{dm}, \tilde{w}^{(2)}_{dm} \mid \tilde{z}_{dm}) \right]  - \sum_{k=1}^{K} E_{q} \left[ \log q (\beta_k \mid \lambda_k) \right] - \sum_{d=1}^{D} E_{q} \left[ \log q (\theta_d \mid \gamma_d) \right] \\ &- \sum_{d=1}^{D} \sum_{n=1}^{N_d} E_{q} \left[ \log q (z_{dn} \mid \phi_{dn}) \right]  - \sum_{d=1}^{D} \sum_{m=1}^{M_d} E_{q} \left[ \log q (\tilde{z}_{dm} \mid \tilde{\phi}_{dm})) \right] 
\end{align*}
Where $\tilde{w}^{(1)}_{dm}$ and $\tilde{w}^{(2)}_{dm} $ denoted as the first word and second word of biterm $\tilde{w}_{dm} $. $\tilde{z}_{dm} $ is the topic assignment of $\tilde{w}_{dm} $.
We can expand this equation as follows:
\begin{align*}
E_{q} \left[ \log P (\beta_k \mid \eta_k) \right] &= \log \Gamma \left( \sum_{v=1}^{V} \eta_{kv} \right) - \sum_{v=1}^{V} \log (\eta_{kv}) + \sum_{v=1}^{V} (\eta_{kv} - 1) E_{q} \left[ \log \beta_{kv} \right] \\
&= \log \Gamma \left( \sum_{v=1}^{V} \eta_{kv} \right) - \sum_{v=1}^{V} \log (\eta_{kv}) + \sum_{v=1}^{V} (\eta_{kv} - 1) \left( \psi (\lambda_{kv}) - \psi \left(\sum_{u=1}^{V} \lambda_{ku} \right) \right) \\ E_{q} \left[ \log P (\theta_d \mid \alpha) \right] &= \log \Gamma \left( \sum_{k=1}^{K} \alpha_{k} \right) - \sum_{k=1}^{K} \log (\alpha_{k}) + \sum_{k=1}^{K} (\alpha_{k} - 1) E_{q} \left[ \log \theta_{dk} \right] \\
&= \log \Gamma \left( \sum_{k=1}^{K} \alpha_{k} \right) - \sum_{k=1}^{K} \log (\alpha_{k}) + \sum_{k=1}^{K} (\alpha_{k} - 1) \left( \psi (\gamma_{dk}) - \psi \left(\sum_{j=1}^{K} \gamma_{dj} \right) \right) \\  E_{q} \left[ \log P (z_{dn} \mid \theta_d) \right] &= \sum_{k=1}^{K} \phi_{dnk} E_{q} [\log \theta_{dk}]  = \sum_{k=1}^{K} \phi_{dnk} \left( \psi (\gamma_{dk}) - \psi \left(\sum_{j=1}^{K} \gamma_{dj} \right) \right) \\  E_{q} \left[ \log P (z_{dn} \mid \theta_d) \right] &= \sum_{k=1}^{K} \phi_{dnk} E_{q} [\log \theta_{dk}]  = \sum_{k=1}^{K} \phi_{dnk} \left( \psi (\gamma_{dk}) - \psi \left(\sum_{j=1}^{K} \gamma_{dj} \right) \right) \\ E_{q} \left[ \log P (\tilde{z}_{dm} \mid \theta_d) \right] &= \sum_{k=1}^{K} \tilde{\phi}_{dmk} E_{q} [\log \theta_{dk}] = \sum_{k=1}^{K} \tilde{\phi}_{dmk} \left( \psi (\gamma_{dk}) - \psi \left(\sum_{j=1}^{K} \gamma_{dj} \right) \right) \\ E_{q} [\log P(w_{dn} \mid \beta_{z_{dn}})] &= \sum_{v=1}^V I\{ w_{dn} = v \} E_{q} [\log \beta_{z_{dn},v} ] =  \sum_{v=1}^V I\{ w_{dn} = v \} \sum_{k=1}^K \phi_{dnk} E_{q} [\log \beta_{kv} ] \\ &= \sum_{v=1}^V \sum_{k=1}^K I\{ w_{dn} = v \} \phi_{dnk} \left( \psi (\lambda_{kv}) - \psi \left(\sum_{u=1}^{V} \lambda_{ku} \right) \right) 
\end{align*}
\begin{align*}
E_{q} \left[ \log P (\tilde{w}^{(1)}_{dm}, \tilde{w}^{(2)}_{dm} \mid \tilde{z}_{dm}) \right] &= E_{q} \left[ \log P (\tilde{w}^{(1)}_{dm} \mid \tilde{z}_{dm}) \right] + E_{q} \left[ \log P (\tilde{w}^{(2)}_{dm} \mid \tilde{z}_{dm}) \right] \\ &= \sum_{v=1}^V \sum_{k=1}^K I\{ \tilde{w}^{(1)}_{dm} = v \} \tilde{\phi}_{dmk} \left( \psi (\lambda_{kv}) - \psi \left(\sum_{u=1}^{V} \lambda_{ku} \right) \right) \\ &+ \sum_{v=1}^V \sum_{k=1}^K I\{ \tilde{w}^{(2)}_{dm} = v \} \tilde{\phi}_{dmk} \left( \psi (\lambda_{kv}) - \psi \left(\sum_{u=1}^{V} \lambda_{ku} \right) \right)
\end{align*}
\begin{align*}
E_{q} \left[ \log P (\beta_k \mid \lambda_k) \right] &= \log \Gamma \left( \sum_{v=1}^{V} \lambda_{kv} \right) - \sum_{v=1}^{V} \log (\lambda_{kv}) + \sum_{v=1}^{V} (\lambda_{kv} - 1) E_{q} \left[ \log \beta_{kv} \right] \\
&= \log \Gamma \left( \sum_{v=1}^{V} \lambda_{kv} \right) - \sum_{v=1}^{V} \log (\lambda_{kv}) + \sum_{v=1}^{V} (\lambda_{kv} - 1) \left( \psi (\lambda_{kv}) - \psi \left(\sum_{u=1}^{V} \lambda_{ku} \right) \right)
\end{align*}
\begin{align*}
E_{q} \left[ \log P (\theta_d \mid \gamma_d) \right] &= \log \Gamma \left( \sum_{k=1}^{K} \gamma_{dk} \right) - \sum_{k=1}^{K} \log (\gamma_{dk}) + \sum_{k=1}^{K} (\gamma_{dk} - 1) E_{q} \left[ \log \theta_{dk} \right] \\
&= \log \Gamma \left( \sum_{k=1}^{K} \gamma_{dk} \right) - \sum_{k=1}^{K} \log (\gamma_{dk}) + \sum_{k=1}^{K} (\gamma_{dk} - 1) \left( \psi (\gamma_{dk}) - \psi \left(\sum_{j=1}^{K} \gamma_{dj} \right) \right) \\ E_{q} \left[ \log P (z_{dn} \mid \phi_{dn}) \right] &= \sum_{k=1}^{K} \phi_{dnk} \log \phi_{dnk} \\ E_{q} \left[ \log P (\tilde{z}_{dm} \mid \tilde{\phi}_{dm}) \right] &= \sum_{k=1}^{K} \tilde{\phi}_{dmk} \log \tilde{\phi}_{dmk}
\end{align*}
Now, we can maximize the lower bound function in each dimension of the variational parameters. 
\subsection{Variation parameter $\lambda$}
Choose a topic index $k$. Fix $\gamma, \phi $ and each $\lambda_j $ for $ j \neq k$. We rewrite the lower bound as follow: 
\begin{align*}
L(\lambda_k) &= \sum_{v=1}^V (\eta_{kv} - 1) \left( \psi(\lambda_{kv}) - \psi \left( \sum_{u=1}^V \lambda_{ku} \right) \right) \\ & + \sum_{d=1}^D \sum_{n=1}^{N_d} \sum_{v=1}^V I\{w_{dn} = v\} \phi_{dnk} \left( \psi(\lambda_{kv}) - \psi \left( \sum_{u=1}^V \lambda_{ku} \right) \right) \\ &+  \sum_{d=1}^D \sum_{m=1}^{M_d} \sum_{b=1}^{\tilde{V}} I\{\tilde{w}_{dm} = b\} \tilde{\phi}_{dmk} \left( \psi(\lambda_{kv}) - \psi \left( \sum_{u=1}^V \lambda_{ku} \right) \right) \\ &- \log \Gamma \left( \sum_{v=1}^V \lambda_{kv} \right) + \sum_{v=1}^V \log \Gamma (\lambda_{kv})- \sum_{v=1}^V (\lambda_{kv} - 1) \left( \psi(\lambda_{kv}) - \psi \left( \sum_{u=1}^V \lambda_{ku} \right) \right) + \text{const}
\end{align*}

The partial derivative of $L(\lambda_k) $ with respect to $\lambda_k $ is:
\begin{align*}
\frac{\partial}{\partial \lambda_{kv}} L(\lambda_{k}) &= - \left( \psi (\lambda_{kv}) - \psi \left( \sum_{u=1}^{V} \lambda_{ku} \right) \right) + \left( \psi_1 (\lambda_{kv}) - \psi_1 \left( \sum_{u=1}^V \lambda_{ku} \right) \right) \\ & . \left( \eta_{kv} - \lambda_{kv} + \sum_{d=1}^{D} \sum_{n=1}^{N_d} I \{ w_{dn} = v \} \phi_{dnk} + \sum_{d=1}^{D} \sum_{m=1}^{M_d} \left( I \{ \tilde{w}_{dm}^{(1)} = v \} + I\{\tilde{w}_{dm}^{(2)} = v\} \right)   \tilde{\phi}_{dmk} \right) \\ &+ \psi (\lambda_{kv}) - \psi \left( \sum_{u=1}^V \lambda_{ku} \right) + \psi_1 \left( \sum_{u=1}^V \lambda_{ku} \right) \\ &  . \left( \eta_{kv} - \lambda_{kv} + \sum_{d=1}^{D} \sum_{n=1}^{N_d} I \{ w_{dn} = v \} \phi_{dnk} + \sum_{d=1}^{D} \sum_{m=1}^{M_d} \left( I \{ \tilde{w}_{dm}^{(1)} = v \} + I\{\tilde{w}_{dm}^{(2)} = v\} \right)   \tilde{\phi}_{dmk} \right) 
\end{align*}

Set the derivative of $L(\lambda_k)$ to zero, we obtain:
\begin{equation}
 \lambda_{k, v} \leftarrow  \eta_{k, v} + \sum_{d=1}^D \sum_{n=1}^{N_d} I\{ \tilde{w}^{(1)}_{dn} = v \} \phi_{dnk}  + \sum_{d=1}^D \sum_{m=1}^{M_d} \left[ I\{ \tilde{w}^{(1)}_{dm} = v \} + I\{ \tilde{w}^{(2)}_{dm} = v \} \right] \tilde{\phi}_{dmk}
\end{equation}

\subsection{Variation parameter $\gamma$}
Choose a document $d$,  fix $\lambda, \phi $ and each $\gamma_c $ for $ c \neq d$. We rewrite the lower bound as follow:
\begin{align*}
L(\gamma_d) &= \sum_{k=1}^{K} (\alpha_k - 1) \left( \psi(\gamma_{dk}) - \psi \left( \sum_{j=1}^K \gamma_{dj} \right) \right) \\ &+ \left( \sum_{n=1}^{N_d} \sum_{k=1}^K \phi_{dnk} + \sum_{m=1}^{M_d} \sum_{k=1}^K \tilde{\phi}_{dmk} \right) \left( \psi(\gamma_{dk}) - \psi \left( \sum_{j=1}^K \gamma_{dj} \right) \right) \\ & - \log \Gamma \left( \sum_{k=1}^K \gamma_{dk} \right) +  \sum_{k=1}^K \log \Gamma \left(\gamma_{dk} \right) +  \sum_{k=1}^K \left(\gamma_{dk} - 1 \right) \left( \psi(\gamma_{dk}) - \psi \left( \sum_{j=1}^K \gamma_{dj} \right) \right) + \text{const} \\ &= \sum_{k=1}^K \left( \alpha_k - \gamma_{dk} + \sum_{n=1}^{N_d}  \phi_{dnk} + \sum_{m=1}^{M_d} \tilde{\phi}_{dmk} \right) \left( \psi(\gamma_{dk}) - \psi \left( \sum_{j=1}^K \gamma_{dj} \right) \right) \\ &- \log \Gamma \left( \sum_{k=1}^K \gamma_{dk} \right) +  \sum_{k=1}^K \log \Gamma \left(\gamma_{dk} \right) + \text{const}
\end{align*}

The partial derivative of $L(\gamma_d) $ with respect to $\gamma_d $ is:
\begin{align*}
\frac{\partial}{\partial \gamma_{dk}} L(\gamma_{d}) &= \psi_1 (\gamma_{dk}) \left( \alpha_k - \gamma_{dk} + \sum_{n=1}^{N_d} \phi_{dnk} + \sum_{m=1}^{M_d} \tilde{\phi}_{dmk} \right) \\ & - \psi_1 \left(\sum_{j=1}^{K} \gamma_{dj} \right) \sum_{j=1}^{K} \left( \alpha_j - \gamma_{dj} + \sum_{n=1}^{N_d} \phi_{dnj} + \sum_{m=1}^{M_d} \tilde{\phi}_{dmj} \right)
\end{align*}

Set the derivative of $L(\gamma_d)$ to zero, we obtain:
\begin{equation}
\gamma_{d, k} = \alpha_{k} + \sum_{n=1}^{N_d} \phi_{dnk} + \sum_{m=1}^{M_d} \tilde{\phi}_{dmk}
\end{equation}
\subsection{Variation parameter $\phi$}
Fixing $\lambda, \gamma, \tilde{\phi} $ and $ \phi_{cu} $ for $(c,u) \neq (d,v) $. We rewrite the lower bound as follow:
\begin{align*}
L(\phi_{dn}) = \sum_{k=1}^K n_{dn} \phi_{dnk} \left( - \log \phi_{dnk} + \psi (\gamma_{dk}) - \psi \left( \sum_{j=1}^K \gamma_{dj} \right) + \psi (\lambda_{kw_{dn}}) - \psi \left( \sum_{u=1}^V \lambda_{ku} \right)   \right) + \text{const}
\end{align*}

The partial derivative of $L(\phi_{dv}) $ with respect to $\phi_{dv} $ is:
\begin{align*}
\frac{\partial}{\partial \phi_{dnk}} L(\phi_{dn}) = n_{dn} \left( - \log \phi_{dnk} + \psi (\gamma_{dk}) - \psi \left( \sum_{j=1}^K \gamma_{dj} \right) + \psi (\lambda_{kw_{dn}}) - \psi \left( \sum_{u=1}^V \lambda_{ku} \right) - 1  \right)
\end{align*}

Using the Lagrange multipliers method with constraint $\sum_{k=1}^K \phi_{dnk} = 1 $, we obtain:
\begin{equation}
\phi_{dnk}  \propto exp \lbrace E_{q} [\log \theta_{d, k}] + E_{q}[\log \beta_{k,w_{dn}}] \rbrace
\end{equation}

\subsection{Variation parameter $\tilde{\phi}$}
Fixing $\lambda, \gamma, \phi $ and $ \tilde{\phi}_{cu} $ for $(c,u) \neq (d,b) $. We rewrite the lower bound as follow:
\begin{align*}
L(\tilde{\phi}_{db}) = \sum_{k=1}^K m_{db} \tilde{\phi}_{d,b,k} \left( - \log \tilde{\phi}_{d,b,k} + \psi (\gamma_{dk}) - \psi \left( \sum_{j=1}^K \gamma_{dj} \right) + \psi (\lambda_{kv}) - \psi \left( \sum_{b=1}^{V_b} \lambda_{kb} \right)   \right) + const
\end{align*}

The partial derivative of $L(\tilde{\phi}_{db}) $ with respect to $\tilde{\phi}_{db} $ is:
\begin{align*}
\frac{\partial}{\partial \tilde{\phi}_{dbk}} L(\tilde{\phi}_{db}) = m_{db} \left( - \log \tilde{\phi}_{d,b,k} + \psi (\gamma_{dk}) - \psi \left( \sum_{j=1}^K \gamma_{dj} \right) + \psi (\lambda_{kv}) - \psi \left( \sum_{b=1}^{V_b} \lambda_{kb} \right) - 1  \right)
\end{align*}

Using the Lagrange multipliers method with constraint $\sum_{k=1}^K \tilde{\phi}_{dmk} = 1 $, we obtain:
\begin{equation}
\tilde{\phi}_{d, m, k}  \propto exp \lbrace E_{q}[\log \theta_{d, k}] + E_{q}[log \beta_{k,w_{dm}^{(1)}}] + E_{q}[\log \beta_{k,w_{dm}^{(2)}}] \rbrace
\end{equation}
Here we donote $\tilde{w}_{dm}^{(1)}$ and $\tilde{w}_{dm}^{(2)}$ as the first word and second word of biterm $\tilde{w}_{dm}$ respectively.
\end{appendices}
\end{document}